\def\cvprPaperID{5608}
\def\confName{CVPR}
\def\confYear{2024}

\def\paperTitle{EvalCrafter: Benchmarking and Evaluating Large Video Generation Models}

\def\authorBlock{
    Yaofang Liu$^{1,2,}$\thanks{Equal Contribution.} \quad
    Xiaodong Cun$^{1,}$\footnotemark[1] \quad
    Xuebo Liu$^{3}$ \quad
    Xintao Wang$^{1}$ \quad \\
    Yong Zhang$^1$ \quad 
    Haoxin Chen$^1$ \quad  
    Yang Liu$^{4}$\thanks{Corresponding Author} \quad
    Tieyong Zeng$^{4}$\quad
    Raymond Chan$^{2}$\footnotemark[2] \quad
    Ying Shan$^{1}$ \quad
    \\
    \textsuperscript{\rm 1}~Tencent AI Lab
    \qquad \textsuperscript{\rm 2}~City University of Hong Kong \qquad  
    \textsuperscript{\rm 3}~University of Macau \qquad \\
    \textsuperscript{\rm 4}~The Chinese University of Hong Kong \\
    {\textbf{Project Page: \url{http://evalcrafter.github.io}}}
}

\newif\ifreview 
\newif\ifarxiv \newcommand{\arxiv}{\arxivtrue}
\newif\ifcamera 
\newif\ifrebuttal 

\arxiv

\pdfoutput=1
\documentclass[10pt,twocolumn]{article}

\ifreview \usepackage[review]{cvpr} \fi
\ifarxiv \usepackage[pagenumbers]{cvpr} \fi
\ifrebuttal \usepackage[rebuttal]{cvpr} \fi
\ifcamera \usepackage{cvpr} \fi

\usepackage{graphicx}
\usepackage{amsmath}
\usepackage{amssymb}
\usepackage{booktabs}

\usepackage{times}
\usepackage{microtype}
\usepackage{epsfig}
\usepackage[table,xcdraw]{xcolor}
\usepackage{caption}
\usepackage{float}
\usepackage{placeins}
\usepackage{color, colortbl}
\usepackage{stfloats}
\usepackage{enumitem}
\usepackage{tabularx}
\usepackage{xstring}
\usepackage{multirow}
\usepackage{xspace}
\usepackage{url}
\usepackage{subcaption}
\usepackage{xcolor}
\usepackage[hang,flushmargin]{footmisc}
\usepackage{titletoc}

\ifcamera \usepackage[accsupp]{axessibility} \fi

% ...

\ifarxiv  \fi

\newcommand{\R}[1]{{%
    \textbf{%
        \ifstrequal{#1}{1}{\textcolor{red}{R#1}}{%
        \ifstrequal{#1}{2}{\textcolor{blue}{R#1}}{%
        \ifstrequal{#1}{3}{\textcolor{magenta}{R#1}}{%
        \ifstrequal{#1}{4}{\textcolor{teal}{R#1}}{%
                           \textcolor{cyan}{R#1}%
        }}}}%
    }%
}}

\usepackage{xr-hyper}

\makeatletter
\newcommand*{\addFileDependency}[1]{
  \typeout{(#1)}
  \@addtofilelist{#1}
  \IfFileExists{#1}{}{\typeout{No file #1.}}
}

\makeatother

\usepackage[toc,page]{appendix}

\usepackage[pagebackref,breaklinks,colorlinks]{hyperref}
\usepackage[capitalize]{cleveref}
\crefname{section}{Sec.}{Secs.}
\crefname{table}{Table}{Tables}
\crefname{figure}{Fig.}{Figs.}

\begin{document}

\definecolor{amber}{rgb}{1.0, 0.75, 0.0}

\hypersetup{
    colorlinks = true,
    linkbordercolor = {white},
    citecolor=blue,
}

\title{\paperTitle}
\author{\authorBlock}

\maketitle
 \begin{abstract}
The vision and language generative models have been overgrown in recent years. For video generation, various open-sourced models and public-available services have been developed to generate high-quality videos. However, these methods often use a few metrics, \eg, FVD~\cite{fvd} or IS~\cite{is}, to evaluate the performance. We argue that it is hard to judge the large conditional generative models from the simple metrics since these models are often trained on very large datasets with multi-aspect abilities. 
Thus, we propose a novel framework and pipeline for exhaustively evaluating the performance of the generated videos. Our approach involves generating a diverse and comprehensive list of 700 prompts for text-to-video generation, which is based on an analysis of real-world user data and generated with the assistance of a large language model.
Then, we evaluate the state-of-the-art video generative models on our carefully designed benchmark, in terms of visual qualities, content qualities, motion qualities, and text-video alignment with 17 well-selected objective metrics. To obtain the final leaderboard of the models, we further fit a series of coefficients to align the objective metrics to the users' opinions. Based on the proposed human alignment method, our final score shows a higher correlation than simply averaging the metrics, showing the effectiveness of the proposed evaluation method. 

\end{abstract}
\section{Introduction}
\label{sec:intro}

The charm of the large generative models is sweeping the world, \eg, the well-known ChatGPT and GPT4~\cite{gpt4} have shown human-level abilities in several aspects, including coding, solving math problems, and even visual understanding, which can be used to interact with our human beings using any knowledge in a conversational way. 
As for the generative models for visual content creation, Stable Diffusion (SD)~\cite{ldm} and SDXL~\cite{sdxl} play very important roles since they are the most powerful publicly available models that can generate high-quality images from any text prompts.

\begin{figure}[tp]
    \centering
    \includegraphics[width=0.8\linewidth]{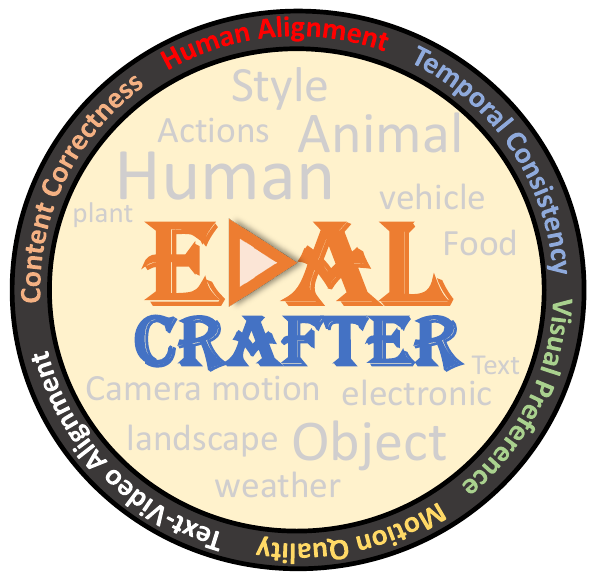}
    \vspace{-1em}
    \caption{We propose EvalCrafter, a comprehensive framework for benchmarking and evaluating the text-to-video models, including the well-defined prompt types in grey and the multiple evaluation aspects in black circles. }
    \vspace{-2em}
    \label{fig:teaser}
\end{figure}

Beyond text-to-image (T2I), taming diffusion model for video generation has also progressed rapidly. Early works~(Imagen-Viedo\cite{imagenvideo}, Make-A-Video~\cite{makeavideo}) utilize the cascaded models for video generation directly. Powered by the image generation priors in SD, LVDM~\cite{lvdm} and MagicVideo~\cite{magicvideo} have been proposed to train the temporal layers to efficiently generate videos. Apart from the academic papers, several commercial services also can generate videos from text or images, \eg, Gen2~\cite{gen1} and PikaLabs~\cite{pika}. Although we can not get the technique details of these services, they are not evaluated and compared with other methods. However, all current large text-to-video (T2V) models only use previous GAN-based metrics like FVD~\cite{fvd} for evaluation, which only concerns the distribution matching between the generated video and the real videos, other than the pairs between the text prompt and the generated video. 
Differently, we argue that a good evaluation method should consider the metrics in different aspects, \eg, the motion quality and the temporal consistency. Also, similar to the large language models (LLMs), some models are not publicly available and we can only get access to the generated videos, which further increases the difficulties in evaluation. Although the evaluation has progressed rapidly in the large generative models, including the areas of LLM~\cite{gpt4}, MLLM~\cite{seedbench}, and T2I~\cite{tifa}, it is still hard to directly use these methods for video generation. The main problem here is that different from T2I or dialogue evaluation, motion and consistency are very important to video generation which previous works ignore. 



We make the very first step to evaluate the general T2V  models. In detail, we first build a comprehensive prompt list containing various everyday objects, attributes, and motions. To achieve a balanced distribution of well-known concepts, we start from the well-defined meta  types of the real-world knowledge and utilize the knowledge of the LLM, \ie, ChatGPT~\cite{gpt4}, to extend our meta-prompt to a wide range. Besides the prompts generated by the model, we also select the prompts from real-world users and T2I prompts. After that, we obtain the metadata~(\eg, color, size, \etc) from the prompt for further evaluation. 
Second, we assess the performance of large T2V models from four aspects, \ie, video quality, text-video alignment, motion quality, and temporal consistency. For each aspect, we employ several objective metrics as evaluation measures, and we conduct a user study to human scores w.r.t. these four aspects. After that, we train coefficients of the regression model for each aspect, aligning evaluation scores with user preferences. 
 This enables us to obtain the final model scores and evaluate new videos using the trained coefficients.

Overall, we summarize the contribution of our paper as:

\begin{itemize}
    \item We make the first step of evaluating the large T2V model and build a comprehensive prompt list with detailed annotations for T2V evaluation.
    \item We consider the aspects of the video visual quality, video motion quality, video temporal consistency, and text-video alignment for the evaluation of video generation. For each aspect, we align the opinions of humans and also verify the effectiveness of the proposed metric by correlation analysis.
    \item During the evaluation, we also discuss several conclusions and findings, which might also contribute to further innovation and development of T2V models.
\end{itemize}

\begin{figure*}[tp]
    \centering
    \includegraphics[width=\linewidth]{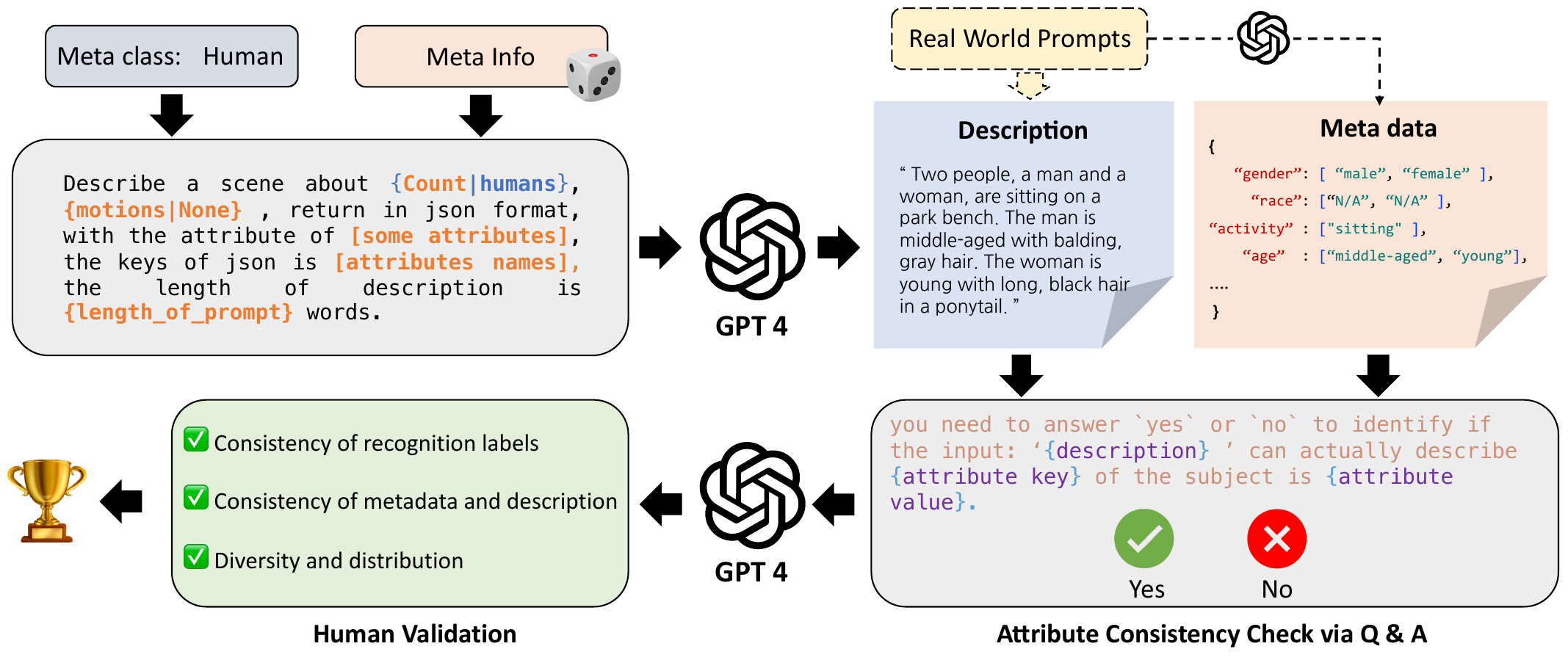}
    \vspace{-2em}
    \caption{We aim to generate a trustworthy benchmark with detailed prompts for text-to-video evaluation by computer vision model and users. We show the pipeline above.}
    \vspace{-1em}
    \label{fig:bench_pipeline}
\end{figure*}

\section{Related Work}
\label{sec:related}

\subsection{Text-to-Video Generation and Evaluation}
Text-to-video (T2V) generation aims to create videos from text prompts. Early works used Variational AutoEncoders (VAEs~\cite{vae}) or generative adversarial networks (GANs~\cite{gan}) but often yielded low-quality or domain-specific results, such as faces~\cite{sadtalker} or landscapes~\cite{stylegan-v, digan}. Recent methods leverage advancements in diffusion models~\cite{ddpm, vdm, vdmsurvey} and large-scale text-image pretraining~\cite{clip} to improve generation quality. Examples include Make-A-Video~\cite{makeavideo}, Imagen-Video~\cite{imagenvideo}, LVDM~\cite{lvdm}, Align Your Latent~\cite{alignlatent}, and MagicVideo~\cite{magicvideo}. Commercial and non-commercial entities have also shown interest in T2V generation, with online services like Gen1~\cite{gen1}, Gen2~\cite{gen1}, and open-source models such as ZeroScope~\cite{zeroscope}, ModelScope~\cite{modelscope}. Discord-based servers like Pika-Lab~\cite{pika} and Morph Studio~\cite{morphstudio} have demonstrated competitive results.

However, a fair and detailed benchmark for evaluating these methods is still lacking. Existing metrics like FVD~\cite{fvd}, IS~\cite{is}, and CLIP similarity~\cite{clip} may perform well on previous in-domain T2I generation methods but do not adequately assess alignment with input text, motion quality, and temporal consistency, which are crucial for T2V.

\subsection{Evaluations on Large Generative Models} 


Evaluating the large generative models~\cite{gpt4, llama, llama2, ldm, sdxl} is a big challenge for both the NLP and vision tasks. For the LLMs, current methods design several metrics in terms of different abilities, question types, and user platform~\cite{tooleval, gu2023xiezhi, hendrycks2021measuring, choi2023llms, zheng2023judging}. More details of LLM evaluation and Multi-model LLM evaluation can be found in recent surveys~\cite{llm-eval, zhao2023survey}. Similarly, the evaluation of the multi-modal generative model also draws the attention of the researchers~\cite{bang2023multitask, ye2023mplug}. For example, Seed-Bench~\cite{seedbench} generates the VQA for multi-modal LLM evaluation. 

For the models in visual generation tasks, Imagen~\cite{imagen} only evaluates the model via user studies. DALL-Eval~\cite{dall-eval} assesses the visual reasoning skills and social basis of the T2I model via both user and object detection algorithm~\cite{detr}.
HRS-Bench~\cite{hrs-bench} proposes a holistic and reliable benchmark by generating the prompt with ChatGPT~\cite{gpt4} and utilizing 17 metrics to evaluate the 13 skills of the T2I model. TIFA~\cite{tifa} proposes a benchmark utilizing the visual question answering~(VQA). However, these methods still work for T2I evaluation or language model evaluation. For T2V evaluation, we further consider the quality of motion and temporal consistency.



\section{Benchmark Construction}
Our benchmark aims to create a trustworthy prompt list to evaluate the abilities of various T2V models fairly. To this end, we first collect and analyze large-scale real-world users' prompts. After that, we propose an automatic pipeline to generate a prompt list with high diversity.
Since video generation is time-consuming, we collect 700 prompts as our initial version for evaluation with careful annotation. 
In this section, we introduce the details of the construction of our benchmark. 

\paragraph{Real-World Data Collection.}
To better understand the types of prompts we should generate, we collect prompts from real-world T2V generation discord users, including the FullJourney~\cite{fulljourney} and PikaLab~\cite{pika}. In total, we gather over 600k prompts with corresponding videos and filter them to 200k by removing repeated and meaningless prompts. 

Through analyzing the collected data including aspects like prompt length and word frequency, we get to know that most of the prompts contain  3 to 40 words. Besides, we identify four meta-subject classes for T2V generation: \texttt{human}, \texttt{animal}, \texttt{object}, and \texttt{landscape}. For each type, we consider the motions and styles of each type, the relationship between the current metaclass and other metaclasses, and the \texttt{motion} and \texttt{camera motion} to construct the benchmark. We give more details in the supplementary materials.

\begin{figure}[tp]
    \centering
    \includegraphics[width=0.9\linewidth]{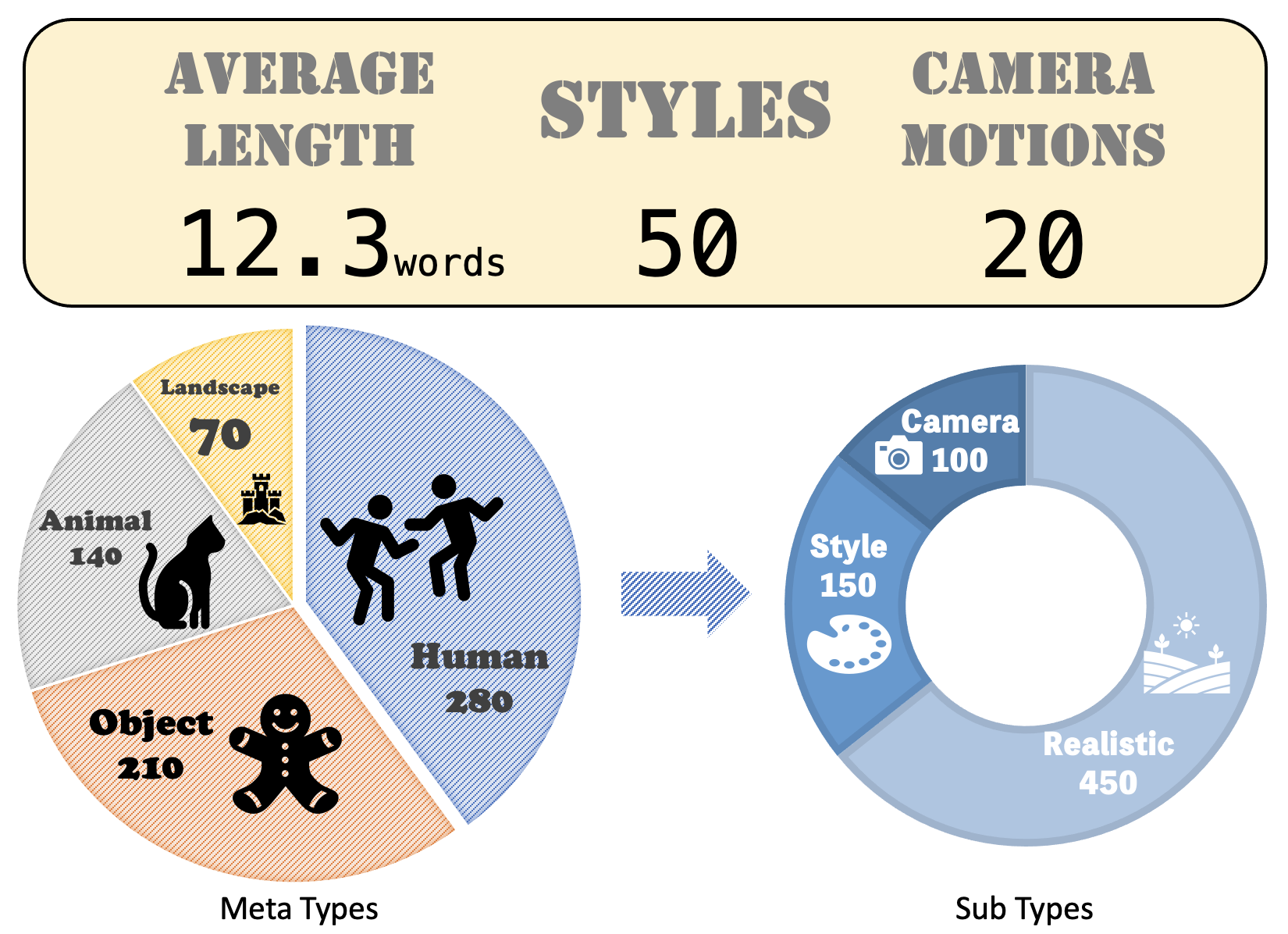}
    \caption{The analysis of the proposed benchmark. Each meta type contains 3 sub-types to increase the generated videos' diversity.}
    \vspace{-1.5em}
    \label{fig:bench_analysis}
\end{figure}

\begin{table*}[t]
\centering
\begin{tabular}{lccccccccc}
\toprule
Method & Ver. & Abilities$^\dag$ & Resolution & FPS & Open Source & Length & Speed$^*$ & Motion & Camera \\
\hline
\rowcolor{lightgray!30}
ModelScope & 23.03 & T2V & 256$\times$256 & 8 & $\checkmark$ & 4s & 0.5 min & - & - \\ 
\rowcolor{lightgray!30}
VideoCrafter & 23.04 & T2V & 256$\times$256 & 8 & $\checkmark$ & 2s & 0.5 min & - & - \\ \hline
ZeroScope & 23.06 & T2V \& V2V & 1024$\times$576  & 8 & $\checkmark$ & 4s & 3 min & - & - \\
ModelScope-XL & 23.08 & I2V \& V2V & 1280$\times$720 & 8 &$\checkmark$ & 4s & 8 min+ & - & - \\ 
Show-1  & 23.10 & T2V & 576$\times$320  & 8 &$\checkmark$ & 4s &  10 min & - & - \\
Hotshot-XL  & 23.10 & T2V & 672$\times$384  & 8 &$\checkmark$ & 1s &  10 s & - & - \\
VideoCrafter1  & 23.10 & I2V \& T2V & 1024$\times$576  & 8 &$\checkmark$ & 2s & 3 min & - & -  \\
Floor33 Pictures & 23.08 & T2V & 1280$\times$720 & 8 & - & 2s & 4 min & - & - \\
PikaLab & 23.09 & I2V \texttt{OR} T2V & 1088$\times$640  & 24 & - & 3s & 1 min & $\checkmark$ & $\checkmark$ \\
Gen2  & 23.09 & I2V \texttt{OR} T2V & 896$\times$512  & 24 & - & 4s & 1 min & $\checkmark$ & $\checkmark$ \\
\bottomrule
\end{tabular}
\vspace{-1em}
\caption{
The difference in the available diffusion-based text-to-video models. $\dag$ We majorly evaluate the method of text-to-video generation~(T2V). For related image-to-video generation model~(I2V), \ie, ModelScope-XL, we first generate the image by Stable Diffusion v2.1 and then perform image-to-video on the generated content.
}
\vspace{-1em}
\label{tab:compare}
\end{table*}

\paragraph{General Recognizable Prompt Generation.} 
Based on the metaclasses identified in the previous step, we generate the recognizable prompts with the help of a LLM and human input. As shown in Fig~\ref{fig:bench_pipeline}, for each kind of metaclass, we ask GPT-4~\cite{gpt4} to describe the scenes about this metaclass and its attributes with randomly sampled meta information. This way, we get the ground truth for the computer vision models for evaluation. However, we find that GPT-4 is not perfect for this task, as the generated attributes are not very consistent with the generated description. Thus, we involve a self-check in the benchmark building process where we use GPT-4 to identify the similarities between the generated description and each metadata. Finally, we filter the prompts by ourselves to ensure each prompt is correct and meaningful for T2V generation.

In addition to the automatically generated prompts, we also integrate prompts from real-world users and available T2I evaluation prompts, such as DALL-Eval~\cite{dall-eval} and Draw-Bench~\cite{imagen}. We filter and generate the metadata using GPT-4, choose suitable prompts with corresponding meta-information as shown in Fig.~\ref{fig:bench_pipeline}, and check the consistency of the meta-information.

\paragraph{Benchmark Overview.}
Overall, we get over 700 prompts in the metaclasses of \texttt{human}, \texttt{animal}, \texttt{objects}, and \texttt{landscape}. Each class contains the natural scenes, the stylized prompts, and the results with explicit camera motion controls. We give a brief view of the benchmark in Fig.~\ref{fig:bench_analysis}.
To increase the diversity of the prompts, our benchmark contains 3 different sub-types, where we have a total of 50 styles and 20 camera motion prompts. We randomly add them in 250 prompts of the whole benchmark. Our benchmark contains an average length of 12.3 words per prompt, which is similar to the real-world prompts we collected.

\section{Evaluation Metrics}
Different from previous FID~\cite{fid} based evaluation metrics, we evaluate the T2V models in different aspects, including the visual quality of the generated video, the text-video alignment, the motion quality, and temporal consistency. Below, we give the detailed metrics.


\subsection{Overall Video Quality Assessment}
We focus on the visual quality of the generated video, which is crucial for user appeal. As distribution-based methods like FVD~\cite{fvd} require ground truth videos, we argue they are unsuitable for general T2V generation cases.

\noindent\textbf{Video Quality Assessment~(VQA$_A$, VQA$_T$).} We employ the Dover~\cite{dover} method to assess generated video quality in terms of aesthetics and technicality. The technical rating measures common distortions like noise and artifacts. Dover~\cite{dover} is trained on a large-scale dataset with labels ranked by real users. We denote the aesthetic and technical scores as VQA$_{A}$ and VQA$_{T}$, respectively.

\noindent\textbf{Inception Score~(IS).} We also use the inception score~\cite{is} as a video quality assessment index, following previous T2V generation papers. The inception score evaluates GAN~\cite{gan} performance using a pre-trained Inception Network~\cite{googlenet} on the ImageNet~\cite{imagenet} dataset. A higher inception score indicates more diverse generated content.

\subsection{Text-Video Alignment}
We evaluate the alignment of input text and generated video in various aspects, including global text prompts, content correctness, and specific attributes. The details of each score are as follows.

\noindent\textbf{Text-Video Consistency~(CLIP-Score).} 
We use the CLIP-Score to quantify the discrepancy between input text prompts and generated videos. Using the pretrained \texttt{ViT-B/32} CLIP model~\cite{clip} as a feature extractor, we obtain frame-wise image embeddings and text embeddings, and compute their cosine similarity. The overall CLIP-Score is then derived by averaging individual scores across all frames. 


\noindent\textbf{Image-Video Consistency~(SD-Score).} 
We propose a new metric, SD-Score, to compare the generated quality with frame-wise SD~\cite{ldm}, considering that most current video diffusion models are fine-tuned on a base SD with a larger scale dataset. Using SDXL~\cite{sdxl}, we generate ${N_1}$ images $\{d_{k}\}_{k=1}^{N_1}$ for every prompt and extract visual embeddings in both generated images and video frames. We calculate the embedding similarity between the generated videos and SDXL images, which helps address the concept forgetting problems when fine-tuning the T2I diffusion model to video models. The final SD-Score is calculated as:
\begin{equation} 
S_{SD} = \frac{1}{M} \sum_{i=1}^M (\frac{1}{N} \sum_{t=1}^{N}(\frac{1}{N_1} \sum_{k=1}^{N_1} \mathcal{C}(emb(x_t^i), emb(d_{k}^{i})))). 
\end{equation}
where $x_t^i$ is the $t$-th frame of the $i$-th video, $\mathcal{C}(\cdot, \cdot)$ is the cosine similarity function, $emb(\cdot)$ means CLIP embedding, $M$ is the total number of testing videos, and $N$ is the total number of frames in each video, where $N_1 = 5$.

\noindent\textbf{Text-Text Consistency~(BLIP-BLEU).} 
We also consider the evaluation between the generated text descriptions of the video and the input prompt. We utilize BLIP2~\cite{li2023blip} for caption generation and use BLEU~\cite{bleu} for evaluation of text alignment:
\begin{equation} 
S_{BB} = \frac{1}{M} \sum_{i=1}^M (\frac{1}{N_2} \sum_{k=1}^{N_2} \mathcal{B}(p^i, {l}_{k}^i)),
\end{equation}
where  $p^i$ is  the $i$-th  prompt,  $\mathcal{B}(\cdot, \cdot)$ is the BLEU similarity scoring function, $\{l^i_k\}_{k=1}^{N_2}$ are BLIP2 generated captions for $i$-th video, and $N_2$ is set to 5 experimentally.

\noindent\textbf{Object and Attributes Consistency~(Detection-Score, Count-Score and Color-Score).}
We employ SAM-Track~\cite{samtrack} to analyze the correctness of the video content. We evaluate T2V models on the existence of objects, as well as the correctness of color and count of objects in text prompts. Specifically, we assess the Detection-Score, Count-Score, and Color-Score as follows:

1. \textit{Detection-Score} ($S_{Det}$): Measures average object presence across videos, calculated as:
\begin{equation} 
   S_{Det}= \frac{1}{M_1}\sum_{i=1}^{M_1}\left(\frac{1}{K}\sum_{k=1}^{K} \sigma^i_{t_k}\right),
\end{equation}
where $M_1$ is the number of prompts with objects, $K$ is the number of frames where detection is performed, and $\sigma^i_{t_k}$ is the detection result for frame $t_k$ in video $i$ (1 if an object is detected, 0 otherwise). In our approach, we perform detection every $I = 5$ frames. Therefore, $K = \left\lceil\frac{N}{I}\right\rceil$.

2. \textit{Count-Score} ($S_{Count}$): Evaluates average object count difference, calculated as:
\begin{equation} 
   S_{Count} = \frac{1}{M_2}\sum_{i=1}^{M_2}\left(1 - \frac{1}{K}\sum_{k=1}^{K} \frac{\left| c^i_{t_k} - \hat{c}^{i} \right|}{\hat{c}^{i}}\right),
\end{equation}
where $M_2$ is the number of prompts with object counts, $c^i_{t_k}$ is the detected object count at frame $t_k$ in video $i$, and $\hat{c}^{i}$ is the ground truth object count for video $i$.

3. \textit{Color-Score} (${S}_{Color} $): Assesses average color accuracy, calculated as:
\begin{equation} 
   {S}_{Color} = \frac{1}{M_3}\sum_{i=1}^{M_3}\left(\frac{1}{K}\sum_{k=1}^{K} s^i_{t_k}\right),
\end{equation}
where $M_3$ is the number of prompts with object colors and $s^i_{t_k}$ is the color accuracy result for frame $t_k$ in video $i$ (1 if the detected color matches the ground truth color, 0 otherwise).

\noindent\textbf{Human Analysis~(Celebrity ID Score).} Human is important for the generated videos as shown in our collected real-world prompts. To this end, we evaluate the correctness of human faces using DeepFace~\cite{deepface}, a popular face analysis toolbox. We calculate the distance between the generated celebrities' faces and real images of the celebrities.
\begin{equation} 
S_{CIS} = \frac{1}{M_4} \sum_{i=1}^{M_4} (\frac{1}{N} \sum_{t=1}^{N}( \min_{k \in \{1, \dots, N_3 \}}\mathcal{D}(x_t^i, f_{k}^{i}))), 
\end{equation}
where $M_4$ is the number of prompts that contain celebrities, $\mathcal{D}(\cdot, \cdot)$ is the Deepface's distance function, $\{f^i_k\}_{k=1}^{N_3}$ are collected celebrities images for $i$-th prompt, and $N_3=3$.

\noindent\textbf{Text Recognition~(OCR-Score)} Another hard case for visual generation is to generate text in the input prompt.
To examine the abilities of current T2V models for text generation, we utilize the widely used toolbox PaddleOCR~\cite{paddleocr} to detect the English text from generated videos. Then, similar to HRS-Bench~\cite{hrs-bench}, we calculate Word Error Rate (WER)~\cite{klakow2002testing}, Normalized Edit Distance (NED)~\cite{sun2019icdar}, Character Error Rate (CER)~\cite{morris2004and}, and get the average.

\subsection{Motion Quality}
For video, we believe the motion quality is a major difference from other domains, such as image. To this end, we consider the quality of motion as one of the main evaluation metrics in our evaluation system. Here, we consider two different motion qualities introduced below. 

\noindent\textbf{Action Recognition~(Action-Score).} For videos about humans, we can easily recognize the common actions via pre-trained models. We use the MMAction2 toolbox~\cite{2020mmaction2} and the pre-trained VideoMAE V2 model~\cite{wang2023videomaev2} to infer human actions in generated videos. We take the classification accuracy as our Action-Score, focusing on Kinetics 400 action classes~\cite{kay2017kinetics}.



\noindent\textbf{Average Flow~(Flow-Score).} We also consider the general motion information of the video. To this end, we use RAFT~\cite{raft}, to extract the dense flows of the video in every two frames. Then, we calculate the average flow on these frames to obtain the average flow score of every specific generated video clip since some methods are likely to generate still videos that are hard to be identified by the temporal consistency metrics.

\noindent\textbf{Amplitude Classification Score~(Motion AC-Score).} Based on the average flow, we further identify whether the motion amplitude in the generated video is consistent with the amplitude specified by the text prompt. To this end, we set an average flow threshold $\rho$ that if surpasses $\rho$, one video will be considered large, and here $\rho$ is set to 5 based on our subjective observation. 

\subsection{Temporal Consistency}
Temporal consistency is also a very valuable field in our generated video. To this end, we involve several metrics for calculation. We list them below.

\noindent\textbf{Warping Error.} We first consider the warping error, which is widely used in previous blind temporal consistency methods~\cite{blind_temporal, fatezero, dvp}. In detail, we first obtain the optical flow of each two frames using the pre-trained optical flow estimation network~\cite{raft}, then, we calculate the pixel-wise differences between the warped image and the predicted image. We calculate the warp differences on every two frames and calculate the final score using the average of all the pairs.

\noindent\textbf{Semantic Consistency~(CLIP-Temp).} Besides pixel-wise error, we also consider the semantic consistency between every two frames, which is also used in previous video editing works~\cite{gen1,fatezero}. Specifically, we consider the cosine similarity of the embeddings of each two consecutive frames $(emb(x_t), emb(x_{t+1}))$ of the generated videos and then get the averages on each two frames.

\noindent\textbf{Face Consistency.} Similar to CLIP-Temp, we evaluate the human identity consistency of the generated videos. Specifically, we select the first frame  $x_1$ as the reference and calculate the cosine similarity of $emb(x_1)$ with $\{emb(x_{t})\}_{t=2}^{N}$. Then, we average the similarities as the final score.

\begin{figure}[tp]
    \centering
    \includegraphics[width=\linewidth]{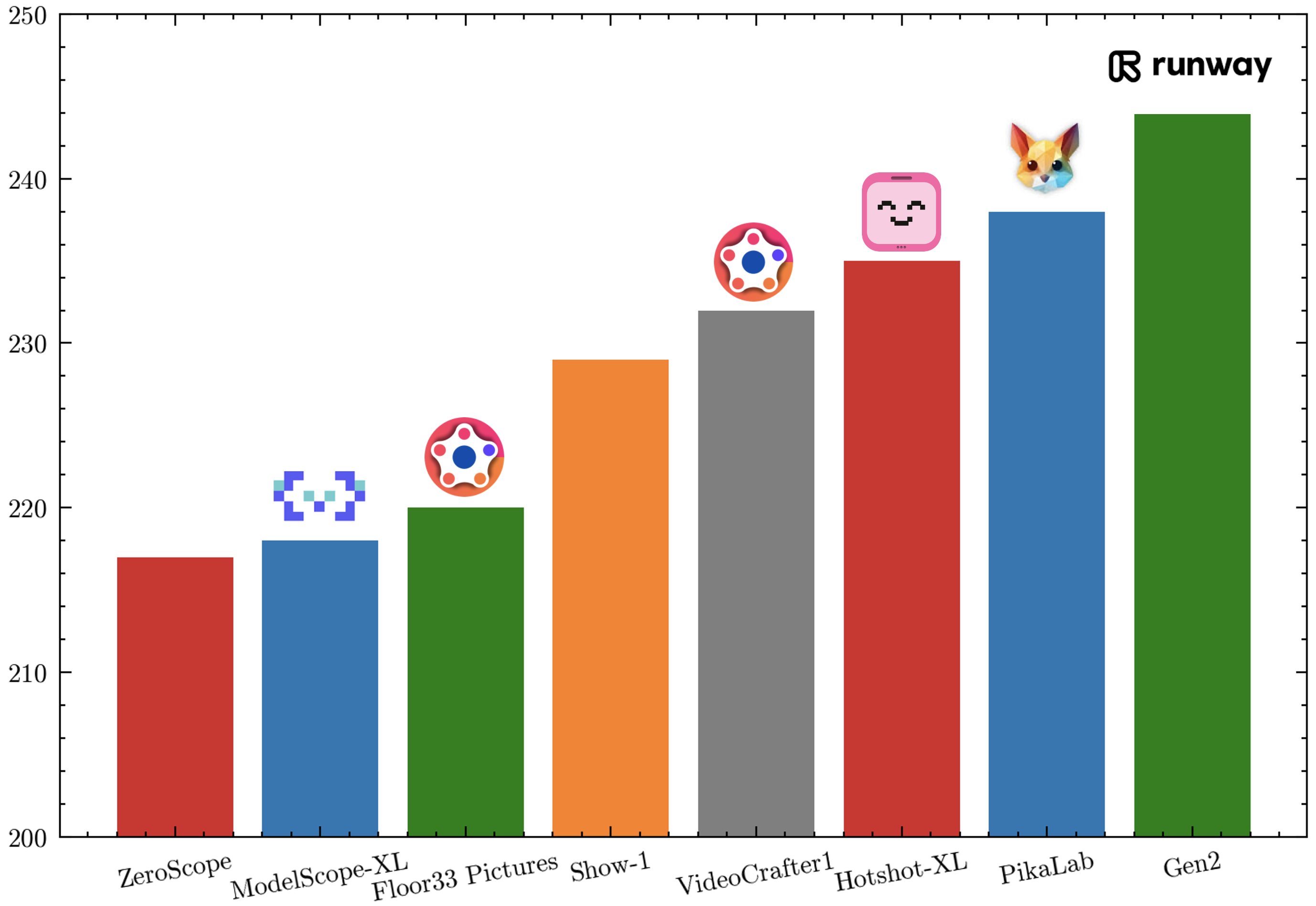}
    \vspace{-2em}
    \caption{Overall comparison results on our EvalCrafter benchmark.
    }
    \vspace{-1em}
    \label{fig:full}
\end{figure}



\begin{table}[t]
\resizebox{\columnwidth}{!}{%
\begin{tabular}{lcccc}
\toprule
{} & Visual & Text-Video & Motion & Temporal \\ 
{} & Quality & Alignment & Quality & Consistency \\ \midrule

ModelScope     & 53.09~(7)   & 54.46~(7)   & 52.47~(7)  & 57.80~(6)      \\ 
ZeroScope      & 53.41~(6)   & 51.21~(8)   & 53.61~(4)  & 58.91~(5)    \\ 
Floor33 Pictures & 58.78~(5) & 61.32~(4)   & 49.16~(8)  & 50.24~(8)  \\ 
PikaLab        & 60.77~(3)   & 55.80~(6)   & 55.77~(2)  & \textbf{65.41~(1)}     \\ 
Gen2           & \textbf{62.51~(1)}   & 60.98~(5)   & \textbf{56.43~(1)}  & 64.41~(2)    \\
VideoCrafter1 & 60.85~(2) & 61.95~(2)   & 53.08~(5) & 55.89~(7) \\
Show-1         & 52.19~(8)   & \textbf{62.07~(1)}   & 53.74~(3) & 60.83~(3) \\
Hotshot-XL        & 60.38~(4)   & 61.52~(3)   & 52.98~(6) & 59.96~(4) \\
\bottomrule
\end{tabular}
}
\vspace{-1em}
\caption{
Human-preference aligned results from four different aspects, with the rank of each aspect in the brackets.
}
\vspace{-1em}
\label{tab:overall comp}
\end{table}
\subsection{User Opinion Alignments}
Besides the above objective metrics, we evaluate user opinions through studies focusing on five main aspects: (1) \textit{Video Quality}, indicating the quality of the generated video where a higher score shows less blur, noise, or other visual degradation; (2) \textit{Text and Video Alignment}, examining the relationships between the generated video and the input text-prompt, requiring users to evaluate the correctness of generated motions; (3) \textit{Motion Quality}, requiring users to identify the correctness of the generated motions from the video.
(4)~\textit{Temporal Consistency}, assessing frame-wise consistency, varying from \textit{Motion Quality}, which needs users to give a rank for high-quality movement;
(5)~\textit{Subjective likeness}, similar to the aesthetic index, a higher value indicates the generated video generally achieves human preference, and we leave this metric used directly. 


\begin{figure}[tp]
    \centering
    \includegraphics[width=\linewidth]{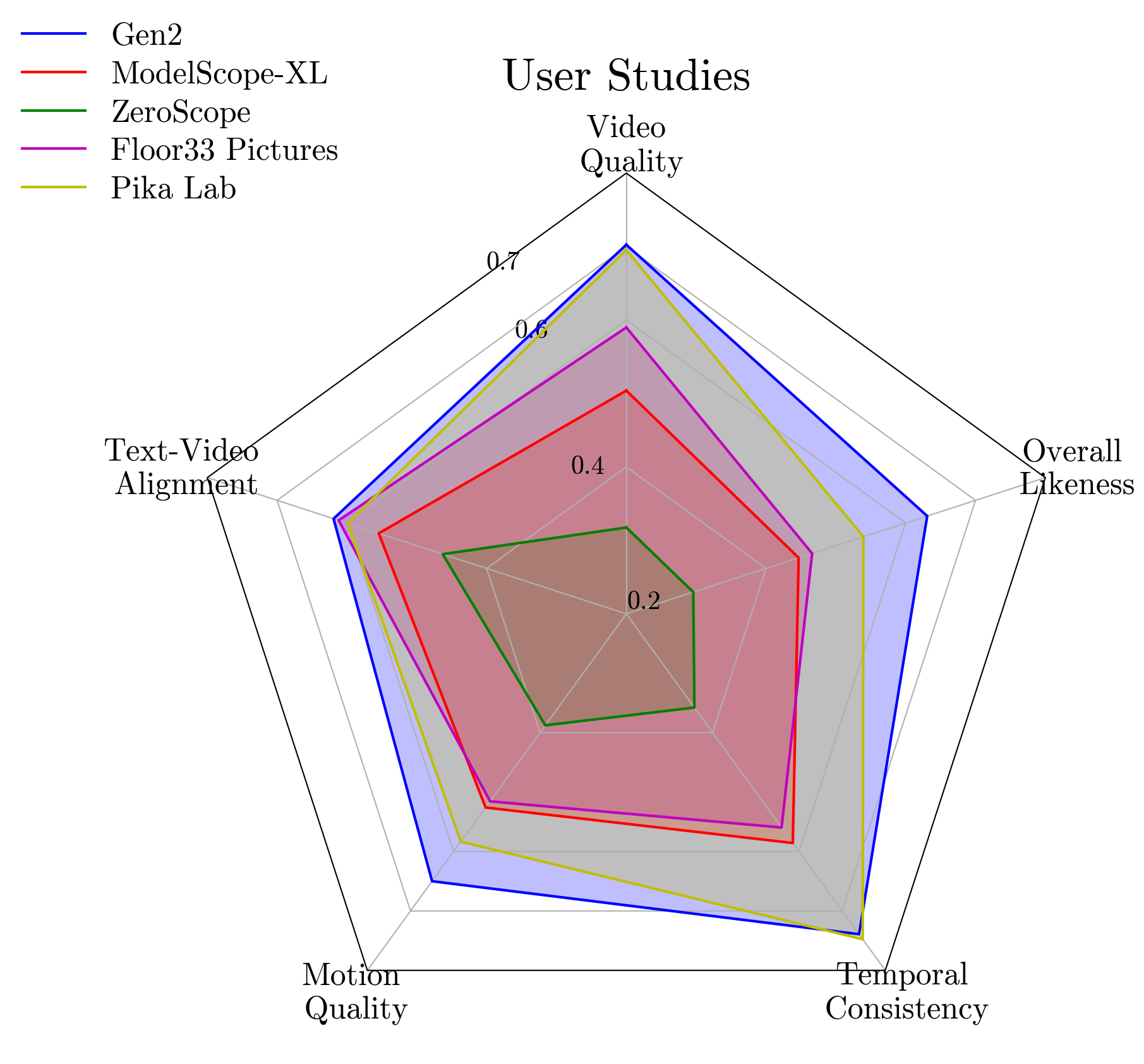}
    \vspace{-2.5em}
    \caption{The raw ratings from our user study. }
    \vspace{-2.0em}
    \label{fig:user_study}
\end{figure}

For evaluation, we generate videos using the provided prompts benchmark on five state-of-the-art methods of ModelScope~\cite{modelscope}, ZeroScope~\cite{zeroscope}, Gen2~\cite{gen1}, Floor33~\cite{floor33}, and PikaLab~\cite{pika}, getting 2.5k videos in total. 
For a fair comparison, we change the aspect ratio of Gen2 and PikaLab to $16:9$ to suitable other methods. Also, since PikaLab can not generate the content without the visual watermark, we add the watermark of PikaLab to all other methods for a fair comparison. 
We also consider that some users might not understand the prompt well, for this purpose, we use SDXL~\cite{sdxl} to generate three reference images of each prompt to help the users understand better, which also inspires us to design an SD-Score to evaluate the models' text-video alignments. For each metric, we ask 7 users to give opinions between 1 to 5, where a large value indicates better alignments. The video sequence has been randomly shuffled before being given to users, and we get 8647 feedback scores in total. Finally, after filtering, we keep 1024 most objective and professional scores as illustrated in Fig.~\ref{fig:user_study}. 

Upon collecting user data, we proceed to perform human alignment for our evaluation metrics, with the goal of establishing a more reliable and robust assessment of T2V algorithms. Initially, we conduct alignment on the data using the mentioned individual metrics above to approximate human scores for the user's opinion in specific aspects.
Similar to the works of the evaluation of natural language processing~\cite{georgila2020predicting, li2021unified}, we employ a linear regression model to fit the parameters in each dimension.
Specifically, we randomly choose 80\% samples from four different methods as the fittings samples and left the rest 20\% samples to verify the effectiveness of the proposed method.
The coefficient parameters are obtained by minimizing the residual sum of squares between the human labels and the prediction from the linear regression model.
In the subsequent stage, we integrate the aligned results of these four aspects and calculate the total score to obtain a comprehensive final score.
\begin{figure*}
     \centering
     \begin{subfigure}[b]{0.33\textwidth}
         \centering
         \includegraphics[width=\textwidth]{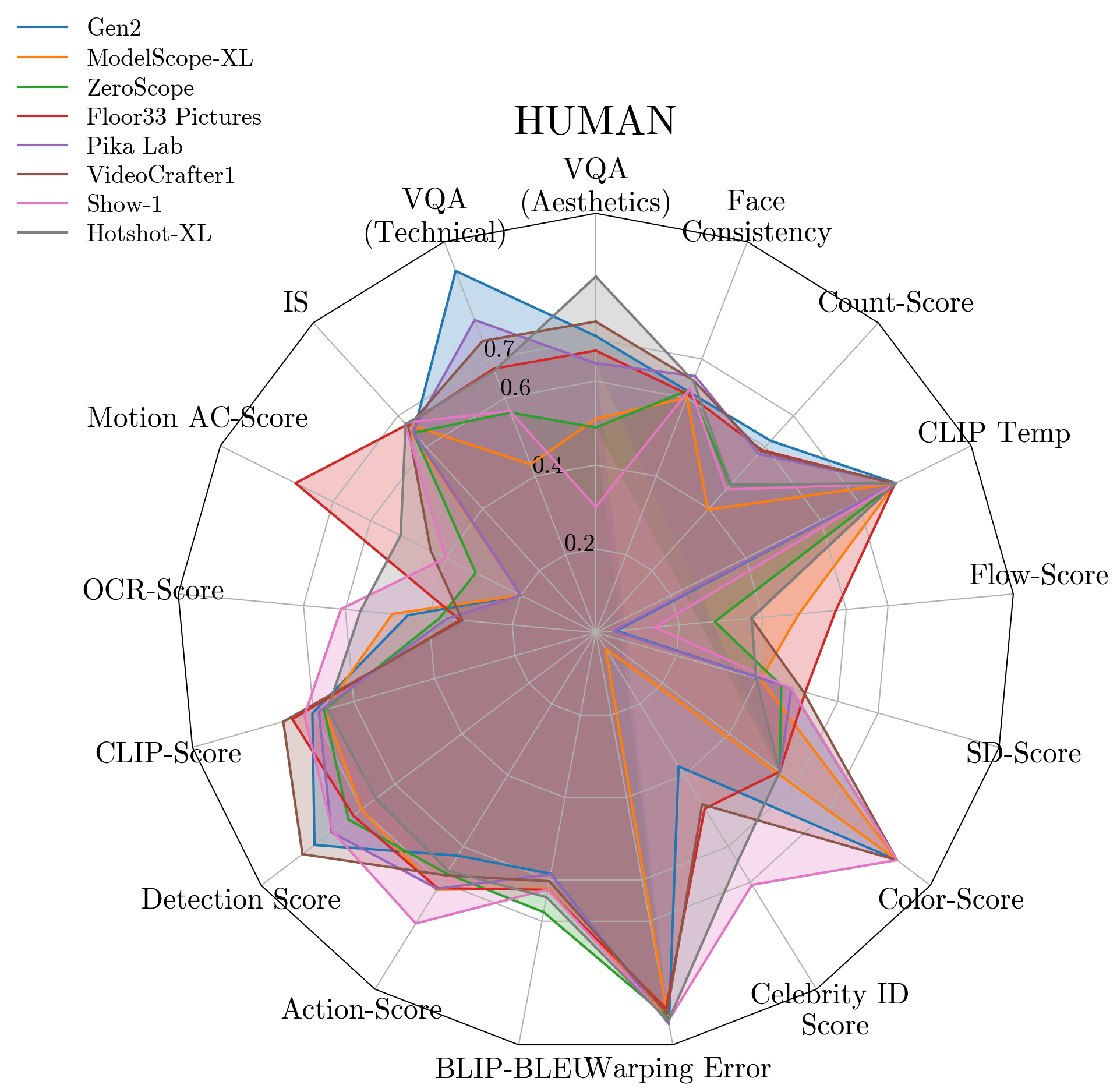}
     \end{subfigure}
     \hfill
     \begin{subfigure}[b]{0.33\textwidth}
         \centering
         \includegraphics[width=\textwidth]{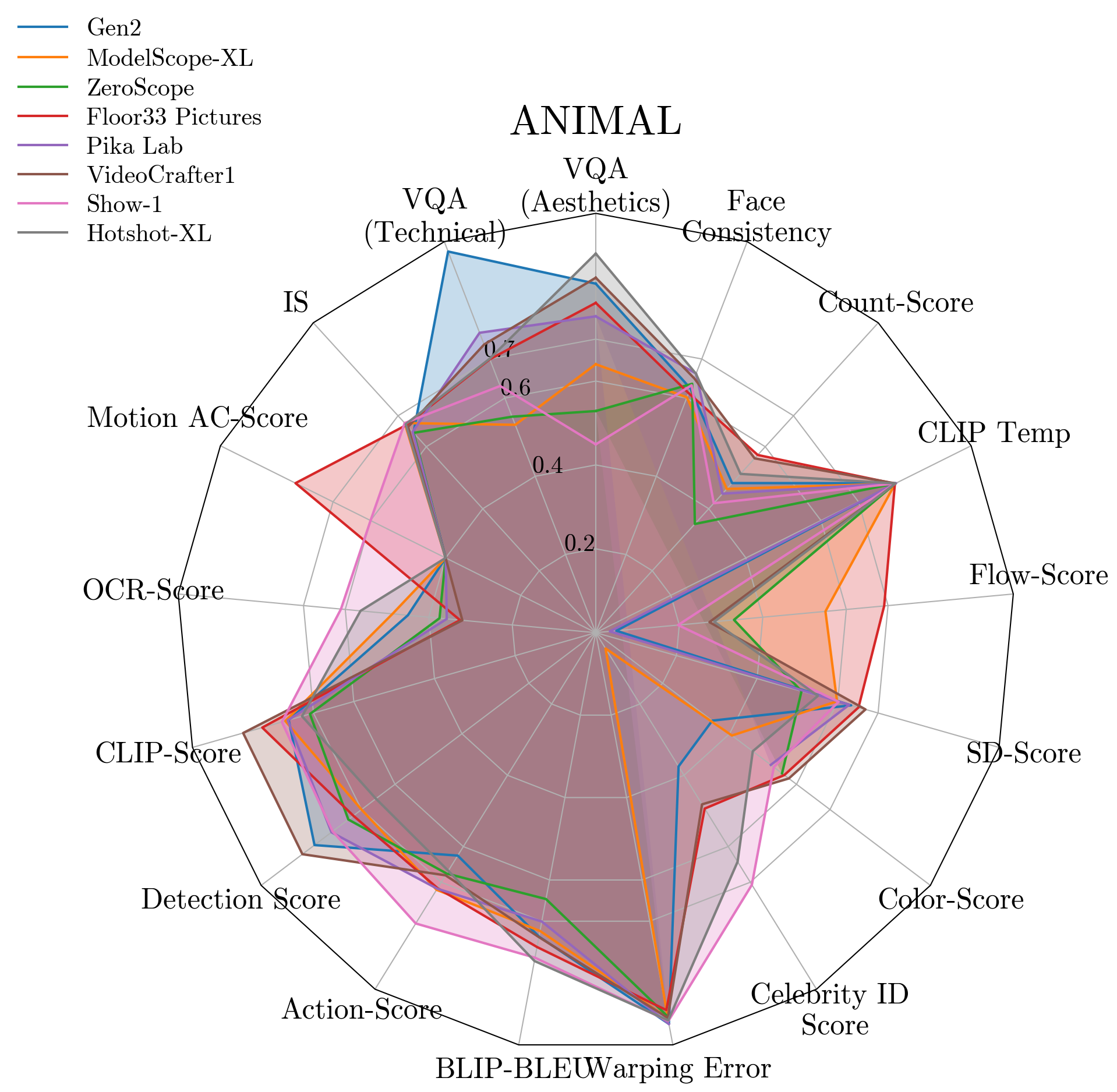}
     \end{subfigure}
     \hfill
     \begin{subfigure}[b]{0.33\textwidth}
         \centering
         \includegraphics[width=\textwidth]{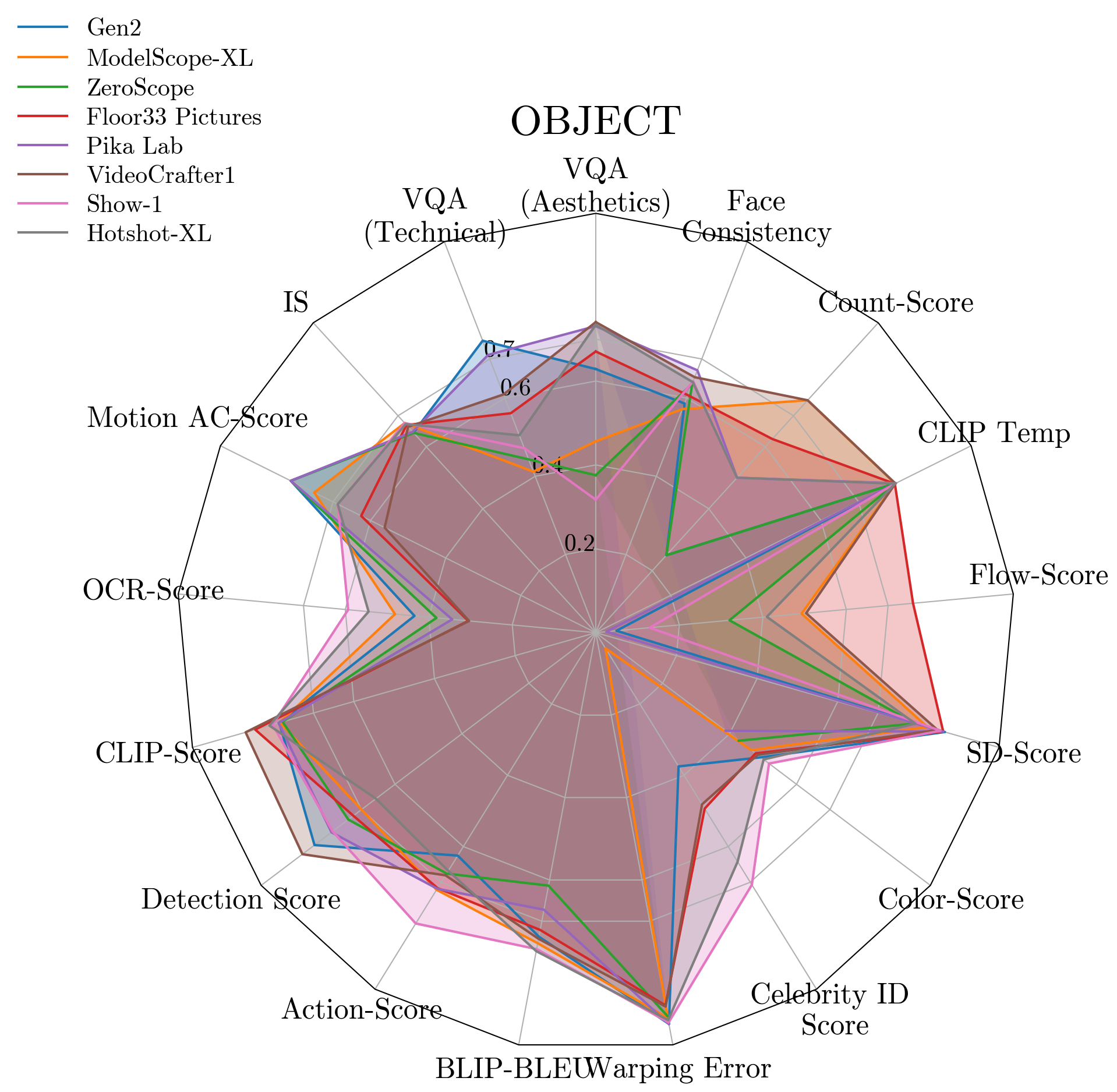}
     \end{subfigure}
     \begin{subfigure}[b]{0.33\textwidth}
         \centering
         \includegraphics[width=\textwidth]{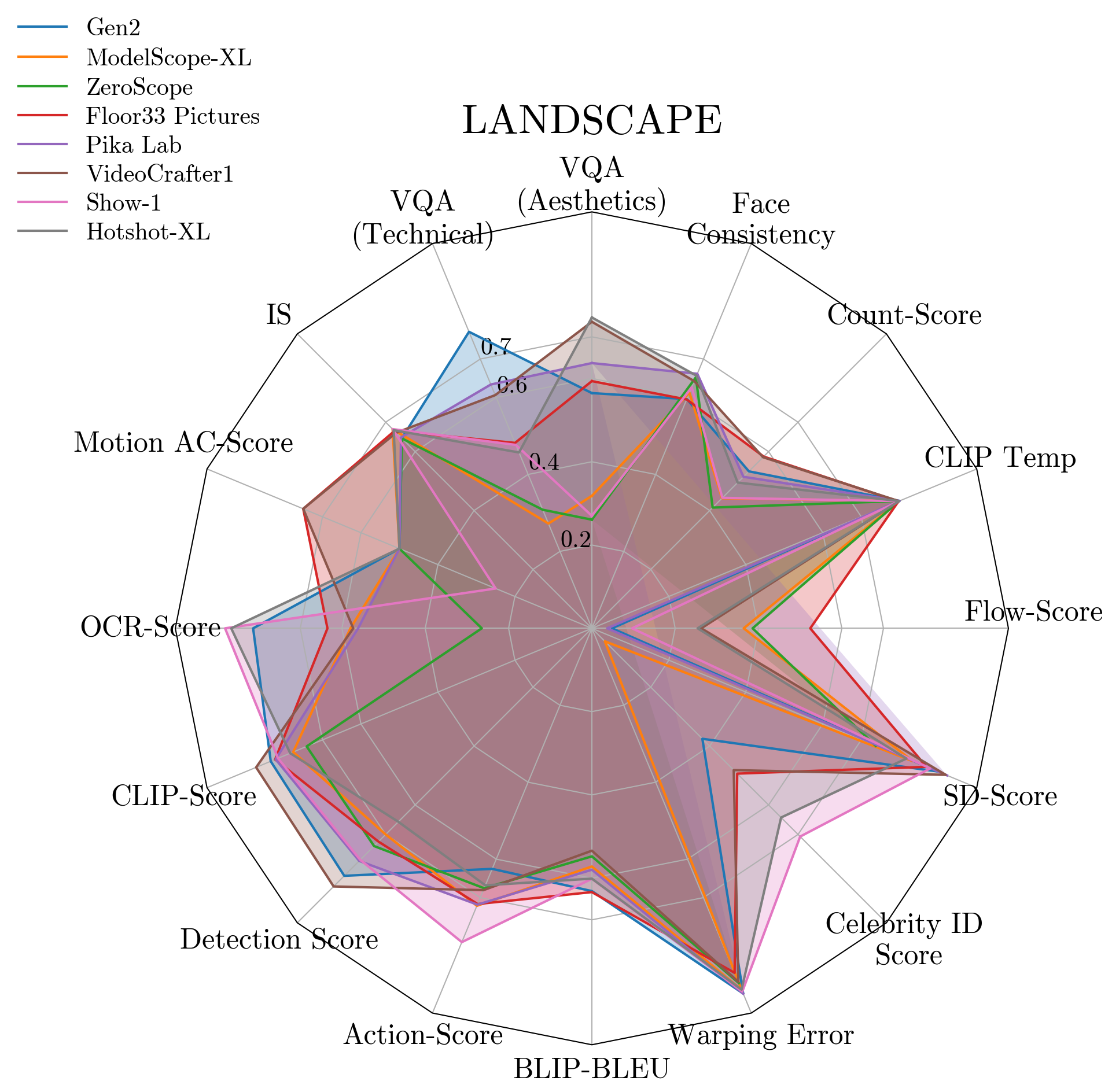}
     \end{subfigure}
     \hfill
     \begin{subfigure}[b]{0.33\textwidth}
         \centering
         \includegraphics[width=\textwidth]{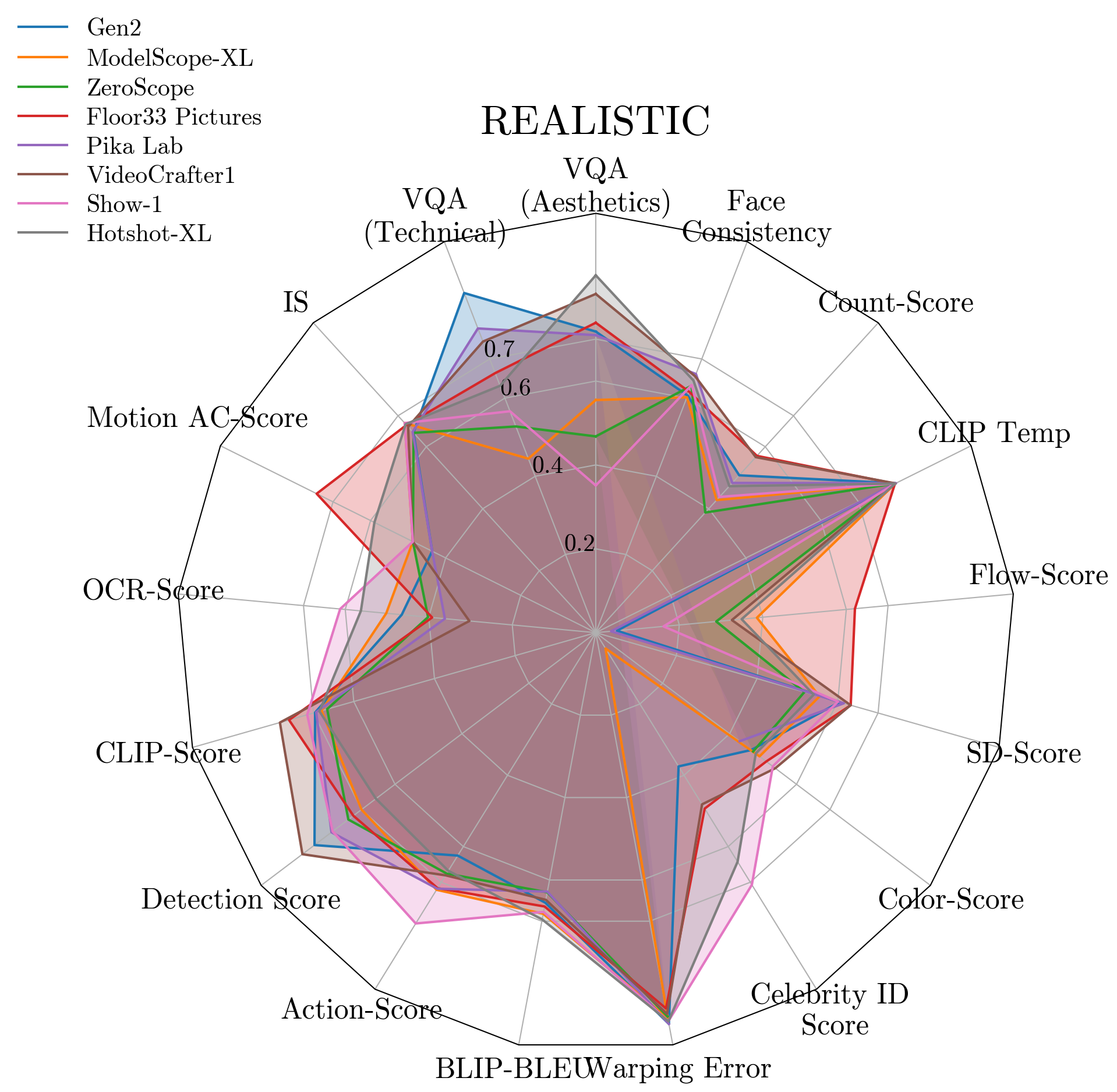}
     \end{subfigure}
     \hfill
     \begin{subfigure}[b]{0.33\textwidth}
         \centering
         \includegraphics[width=\textwidth]{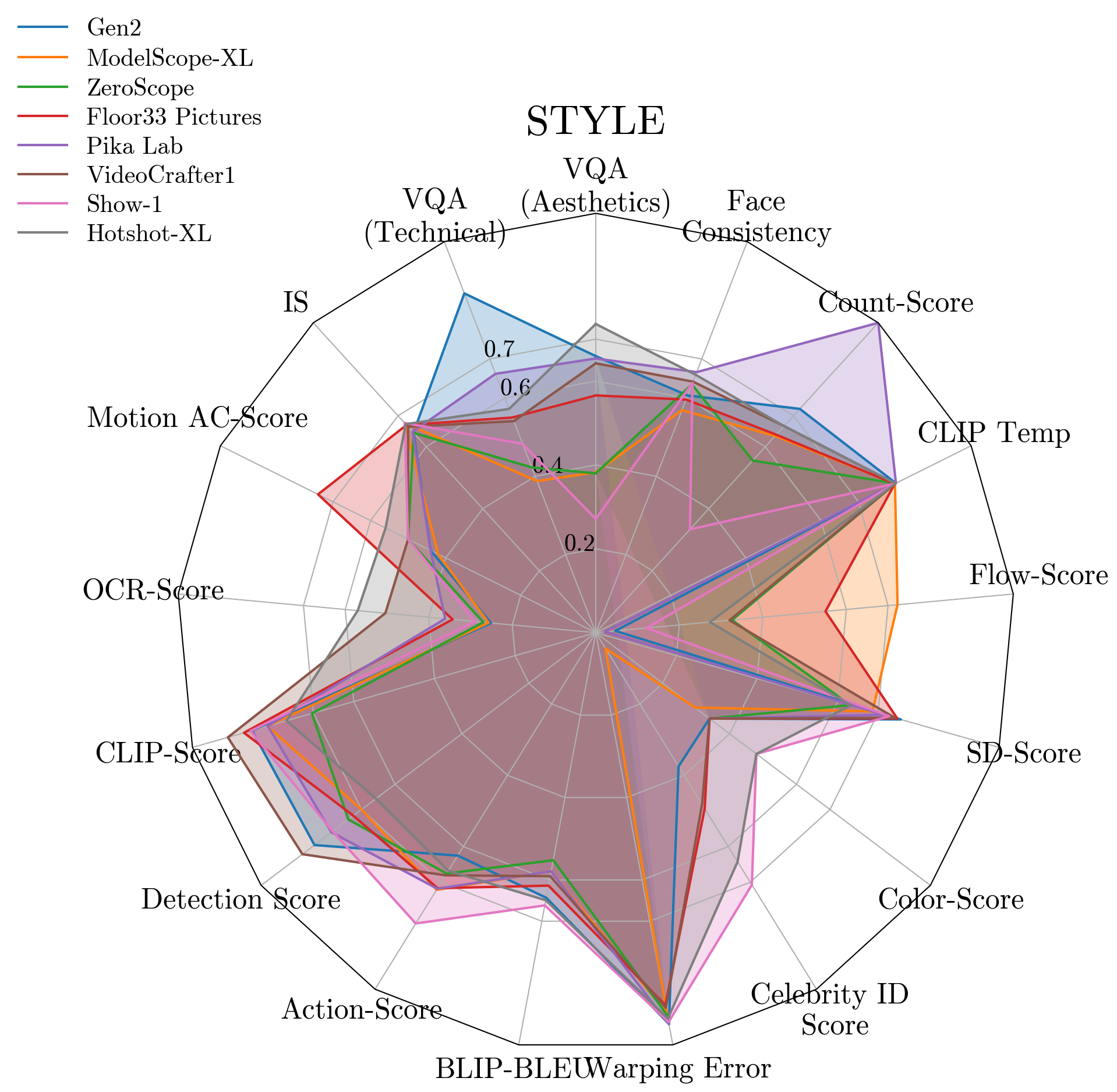}
     \end{subfigure}
     \vspace{-1em}
        \caption{\textbf{\textit{Raw results in different aspects.}} We consider 4 main meta types~(\texttt{animal}, \texttt{human}, \texttt{landscape}, \texttt{object}) to evaluate the performance of the meta types of the generated video, where each type contains several prompts with fine-grained attribute labels. For each prompt, we also consider the style of the video, yet more diverse prompts. as shown in \texttt{realistic} and \texttt{style} figure above. (The metrics values are normalized for better visualization, the Warping Error, Celebrity ID Score, and OCR-Score by $1-$ so that large values indicate better performance.) }
        \vspace{-1em}
        \label{fig:meta_results}
\end{figure*}

\section{Results}
We conduct the evaluation on our benchmark prompts, where each prompt has a metafile for additional information as the answer of evaluation. 
We then generate the videos using all available high-resolution T2V models, including the ModelScope~\cite{modelscope}, Floor33 Pictures~\cite{floor33}, and ZeroScope~\cite{zeroscope}, Show-1~\cite{zhang2023show},
Hotshot-XL~\cite{Hotshot-XL}, and VideoCrafter1~\cite{chen2023videocrafter1}. We keep all the hyper-parameters, such as classifier-free guidance, as the default value. For the service-based model, we evaluate the performance of the representative works of Gen2~\cite{gen1} and PikaLab~\cite{pika}.
They generate at least 512p videos with high-quality watermark-free videos. 
We run all available models on an NVIDIA A100 for speed comparison. 
We first show the overall human-aligned results in Fig.~\ref{fig:full}, with also the different aspects of our benchmark in Table~\ref{tab:overall comp}, which gives us the final and the main metrics of our benchmark. 
Finally, as in Fig.~\ref{fig:meta_results}, we give the results of each method on 4 different meta types~(\ie, \texttt{animal}, \texttt{human}, \texttt{landscape}, \texttt{object}) and two different types of videos~(\ie, \texttt{realistic}, \texttt{style}) in our benchmark. 





\subsection{Analysis on Human Preference Alignment}
To demonstrate the effectiveness of our model in aligning with human scores, we calculate Spearman's rank correlation coefficient~\cite{zar2005spearman} and Kendall's rank correlation coefficient~\cite{kendall1948rank}, both of which are non-parametric measures of rank correlation. 
These coefficients provide insights into the magnitude and direction of the association between our method results and human scores, as listed in Table.~\ref{tab:correlations}. 
From this table, the proposed weighting method shows a better correlation on the unseen 20\% samples than directly averaging.

\subsection{Findings}
\label{sec:analy}
\noindent\textbf{Finding \#1: Single dimension evaluation is insufficient for nowadays T2V models.} Models' rankings in Table.~\ref{tab:overall comp} vary significantly across different aspects, emphasizing the importance of a multi-aspect evaluation approach for a comprehensive understanding of model performance. 

\noindent\textbf{Finding \#2: Meta type evaluation is necessary.} As shown in Fig.~\ref{fig:meta_results}, models perform differently in various meta types,  highlighting the importance of evaluating their abilities by meta type. For example, Gen2~\cite{gen1} behaves better than Floor33 Pictures~\cite{floor33} w.r.t. VQA$_A$ in \texttt{human}, \texttt{animal}, and \texttt{style} videos. Contrarily, it falls behind Floor33 Pictures in \texttt{landscape}, \texttt{object}, and \texttt{realistic} ones.

\noindent\textbf{Finding \#3: Users prioritize visual appeal over T2V alignment.} As shown in \ref{fig:user_study}, despite Gen2~\cite{gen1} performing relatively badly in T2V alignment, it surpasses all other models in \texttt{Subjective Likeness}. We argue that it is because users prefer videos with better visual appeal like good visual quality and high temporal consistency. 

\noindent\textbf{Finding \#4: All methods cannot perform camera motion control directly using text prompt.}  Although some additional hyper-parameters can be set as additional control handles for Gen2~\cite{gen1} and PikaLab~\cite{pika},  all current T2V models still lack the understanding of open-world prompts, like camera motion.


\noindent\textbf{Finding \#5: Resolution doesn't correlate much with visual appeal.} As shown in Table.~\ref{tab:compare} and Table.~\ref{tab:overall comp}, Gen2~\cite{gen1} and  Hotshot-XL~\cite{Hotshot-XL} have small resolutions but are both competitive in visual quality. 
 
\noindent\textbf{Finding \#6:  Larger motion amplitude doesn't ensure user preference.} In our study, most videos that users are fond of are with slight movements, such as those videos generated by PikaLab~\cite{pika} and Gen2~\cite{gen1}. 

\noindent\textbf{Finding \#7: Generating text remains challenging.} Most methods struggle to generate high-quality and consistent text from prompts, as evident from OCR-Scores. Raw results of all metrics are given in supplementary materials.


\noindent\textbf{Finding \#8: Many models can sometimes generate completely wrong videos.} From our study, we find quite a number of failure cases like severe noises and distortion from our baseline models such as ZeroScope~\cite{zeroscope}, ModelScope~\cite{modelscope} and Floor33 Pictures~\cite{floor33}. We argue that it could be viewed as a catastrophic forgetting problem~\cite{shao2022overcoming}, as we know many current T2V models are finetuned from base models like SD~\cite{ldm}. We present our detailed qualitative results in supplementary materials.

\noindent\textbf{Finding \#9: Effective metrics and not that effective metrics.} Metrics like Warp Error, CLIP-Temp, VQA$_T$, and VQA$_A$ seem to perform well as they all have high correlations with human scores shown in Table. \ref{tab:correlations}. However, some metrics are not as good as 
 we think. The  Clip-Score especially, which is a widely used metric in previous works~\cite{gen1,imagenvideo,makeavideo}, only has Spearsman's  $\rho$  6.3 and Kendall's $\phi$  4.3 compared to BLIP-BLEU in the same aspects has 26.7 and 19.0. Detailed correlation results can be found in the supplementary materials.

\noindent\textbf{Finding \#10: All current models are not satisfactory enough.} From our objective evaluation and subjective observation, we argue that T2V models nowadays still have lots to improve. Even for the best model in our evaluation, Gen2~\cite{gen1} also has limitations like struggling with complex scenes, instruction following, and entity details.

\begin{table}[t]
\centering

\begin{tabular}{cl|cc}
\toprule
\multirow{2}{*}{Aspects} & \multirow{2}{*}{Methods} & Spearsman's & Kendall's  \\
 &  & $\rho$ & $\phi$ \\
\hline
\multirow{4}{*}{\shortstack{Visual\\Quality}}   & VQA$_A$ & 42.1  &  30.5    \\
                                  & VQA$_T$ & 53.6  &  39.1    \\
                                  & Avg.    & 55.0  &  41.0    \\
                                  & \textbf{Ours}    & \textbf{55.4}  &  \textbf{41.1}    \\ \hline\hline

\multirow{4}{*}{\shortstack{Motion\\Amplitude}}       
                                  & Motion AC & -22.1  & -16.4    \\
                                  & Flow-Score & -43.3  & -30.1    \\
                                  & Avg.     & -38.2  & -27.7    \\
                                  &\textbf{Ours}       & \textbf{45.0}  &  \textbf{32.4}    \\ \hline\hline

\multirow{4}{*}{\shortstack{Temporal\\Consistency}} & CLIP-Temp & 49.8  & 35.7    \\
                          & Warping Error & 69.0  & 51.7   \\
                          & Avg.     &   54.4  & 38.9    \\
                          & \textbf{Ours}       & \textbf{56.7}  &  \textbf{41.5}  \\ \hline\hline

\multirow{4}{*}{\shortstack{TV\\Alignment}} 
                        & CLIP-Score   & 6.3  & 4.3  \\
                          &  BLIP-BLEU   & 26.7  & 19.0  \\
                          & Avg.         & 31.9 & \textbf{22.7} \\
                          & \textbf{Ours}  & \textbf{32.3}  &  22.5  \\

\bottomrule
\end{tabular}
\vspace{-1em}
\caption{\textbf{\textit{Correlation Analysis.}}
Correlations between some objective metrics and human judgment on text-to-video generations. We use Spearsman's $\rho$ and Kendall's $\phi$ for correlation calculation.
}
\vspace{-1.5em}
\label{tab:correlations}
\end{table}

\subsection{Limitation}
Although we have already made a step in evaluating the T2V generation, there are still many challenges. \textit{(i)}~Currently, we only collect 700 prompts as the benchmark, where the real-world situation is very complicated. More prompts will show a more general benchmark. \textit{(ii)}~Evaluating the motion quality of the general senses is also hard. However, in the era of multi-model LLM and large video foundational models, we believe better and larger video understanding models will be released and we can use them as our metrics. 
\textit{(iii)}~The labels used for alignment are collected from only fewer human annotators, which may introduce some bias in the results. To address this limitation, we plan to expand the pool of annotators and collect more diverse scores to ensure a more accurate and unbiased evaluation.

\section{Conclusion}
\label{sec:conclusion}

Exploring the capabilities of large generative models is crucial for improving model design and utilization. In this paper, we take the first step towards evaluating large, high-quality T2V models by constructing a comprehensive prompt benchmark for T2V assessment. We also provide several objective evaluation metrics to measure T2V model performance concerning video quality, text-video alignment, temporal consistency, and motion quality. Furthermore, we conduct human alignment to correlate user scores with objective metrics, resulting in accurate evaluation metrics for T2V methods. Our experiments demonstrate that the proposed methods effectively align with user opinions, thus providing a reliable assessment of T2V approaches.
We believe this comprehensive evaluation benchmark will serve as a foundation and foster development for future research.

\section*{Acknowledgments}
\label{sec:ack}
Xuebo Liu was sponsored by CCF-Tencent Rhino-Bird Open Research Fund. We would like to thank the anonymous reviewers and meta-reviewer for their insightful suggestions.


{\small
\bibliographystyle{ieee_fullname}
\bibliography{11_references}

\begin{thebibliography}{10}\itemsep=-1pt

\bibitem{floor33}
Floor33 pictures discord server.
\newblock \url{https://www.morphstudio.com/}.
\newblock Accessed: 2023-08-30.

\bibitem{fulljourney}
Fulljourney discord server.
\newblock \url{https://www.fulljourney.ai/}.
\newblock Accessed: 2023-08-30.

\bibitem{Hotshot-XL}
Hotshot-xl.
\newblock \url{https://huggingface.co/hotshotco/Hotshot-XL}.
\newblock Accessed: 2023-10-11.

\bibitem{morphstudio}
Morph studio discord server.
\newblock \url{https://www.morphstudio.com/}.
\newblock Accessed: 2023-08-30.

\bibitem{pika}
Pika {L}ab discord server.
\newblock \url{https://www.pika.art/}.
\newblock Accessed: 2023-08-30.

\bibitem{zeroscope}
Zeroscope.
\newblock \url{https://huggingface.co/cerspense/zeroscope_v2_576w}.
\newblock Accessed: 2023-08-30.

\bibitem{hrs-bench}
EslamMohamed Bakr, Pengzhan Sun, Xiaoqian Shen, FaizanFarooq Khan, LiErran Li,
  and Mohamed Elhoseiny.
\newblock Hrs-bench: Holistic, reliable and scalable benchmark for
  text-to-image models.
\newblock Apr 2023.

\bibitem{bang2023multitask}
Yejin Bang, Samuel Cahyawijaya, Nayeon Lee, Wenliang Dai, Dan Su, Bryan Wilie,
  Holy Lovenia, Ziwei Ji, Tiezheng Yu, Willy Chung, et~al.
\newblock A multitask, multilingual, multimodal evaluation of chatgpt on
  reasoning, hallucination, and interactivity.
\newblock {\em arXiv preprint arXiv:2302.04023}, 2023.

\bibitem{alignlatent}
Andreas Blattmann, Robin Rombach, Huan Ling, Tim Dockhorn, Seung~Wook Kim,
  Sanja Fidler, and Karsten Kreis.
\newblock Align your latents: High-resolution video synthesis with latent
  diffusion models.
\newblock In {\em IEEE Conference on Computer Vision and Pattern Recognition
  ({CVPR})}, 2023.

\bibitem{detr}
Nicolas Carion, Francisco Massa, Gabriel Synnaeve, Nicolas Usunier, Alexander
  Kirillov, and Sergey Zagoruyko.
\newblock End-to-end object detection with transformers.
\newblock In {\em European conference on computer vision}, pages 213--229.
  Springer, 2020.

\bibitem{llm-eval}
Yupeng Chang, Xu Wang, Jindong Wang, Yuan Wu, Kaijie Zhu, Hao Chen, Linyi Yang,
  Xiaoyuan Yi, Cunxiang Wang, Yidong Wang, et~al.
\newblock A survey on evaluation of large language models.
\newblock {\em arXiv preprint arXiv:2307.03109}, 2023.

\bibitem{chen2023videocrafter1}
Haoxin Chen, Menghan Xia, Yingqing He, Yong Zhang, Xiaodong Cun, Shaoshu Yang,
  Jinbo Xing, Yaofang Liu, Qifeng Chen, Xintao Wang, et~al.
\newblock Videocrafter1: Open diffusion models for high-quality video
  generation.
\newblock {\em arXiv preprint arXiv:2310.19512}, 2023.

\bibitem{samtrack}
Yangming Cheng, Liulei Li, Yuanyou Xu, Xiaodi Li, Zongxin Yang, Wenguan Wang,
  and Yi Yang.
\newblock Segment and track anything.
\newblock {\em arXiv preprint arXiv:2305.06558}, 2023.

\bibitem{dall-eval}
Jaemin Cho, Abhay Zala, and Mohit Bansal.
\newblock Dall-eval: Probing the reasoning skills and social biases of
  text-to-image generative transformers.

\bibitem{choi2023llms}
Minje Choi, Jiaxin Pei, Sagar Kumar, Chang Shu, and David Jurgens.
\newblock Do llms understand social knowledge? evaluating the sociability of
  large language models with socket benchmark.
\newblock {\em arXiv preprint arXiv:2305.14938}, 2023.

\bibitem{2020mmaction2}
MMAction2 Contributors.
\newblock Openmmlab's next generation video understanding toolbox and
  benchmark.
\newblock \url{https://github.com/open-mmlab/mmaction2}, 2020.

\bibitem{imagenet}
Jia Deng, Wei Dong, Richard Socher, Li-Jia Li, Kai Li, and Li Fei-Fei.
\newblock Imagenet: A large-scale hierarchical image database.
\newblock In {\em 2009 IEEE Conference on Computer Vision and Pattern
  Recognition}, pages 248--255, 2009.

\bibitem{gen1}
Patrick Esser, Johnathan Chiu, Parmida Atighehchian, Jonathan Granskog, and
  Anastasis Germanidis.
\newblock Structure and content-guided video synthesis with diffusion models.
\newblock {\em arXiv preprint arXiv:2302.03011}, 2023.

\bibitem{georgila2020predicting}
Kallirroi Georgila, Carla Gordon, Volodymyr Yanov, and David Traum.
\newblock Predicting ratings of real dialogue participants from artificial data
  and ratings of human dialogue observers.
\newblock In {\em Proceedings of the Twelfth Language Resources and Evaluation
  Conference}, pages 726--734, 2020.

\bibitem{gan}
Ian Goodfellow, Jean Pouget-Abadie, Mehdi Mirza, Bing Xu, David Warde-Farley,
  Sherjil Ozair, Aaron Courville, and Yoshua Bengio.
\newblock Generative adversarial nets.
\newblock {\em Advances in neural information processing systems}, 27, 2014.

\bibitem{lvdm}
Yingqing He, Tianyu Yang, Yong Zhang, Ying Shan, and Qifeng Chen.
\newblock Latent video diffusion models for high-fidelity long video
  generation.
\newblock 2022.

\bibitem{hendrycks2021measuring}
Dan Hendrycks, Collin Burns, Saurav Kadavath, Akul Arora, Steven Basart, Eric
  Tang, Dawn Song, and Jacob Steinhardt.
\newblock Measuring mathematical problem solving with the math dataset.
\newblock {\em arXiv preprint arXiv:2103.03874}, 2021.

\bibitem{imagenvideo}
Jonathan Ho, William Chan, Chitwan Saharia, Jay Whang, Ruiqi Gao, Alexey
  Gritsenko, Diederik~P Kingma, Ben Poole, Mohammad Norouzi, David~J Fleet,
  et~al.
\newblock Imagen video: High definition video generation with diffusion models.
\newblock {\em arXiv preprint arXiv:2210.02303}, 2022.

\bibitem{ddpm}
Jonathan Ho, Ajay Jain, and Pieter Abbeel.
\newblock Denoising diffusion probabilistic models.
\newblock {\em Advances in neural information processing systems},
  33:6840--6851, 2020.

\bibitem{vdm}
Jonathan Ho, Tim Salimans, Alexey Gritsenko, William Chan, Mohammad Norouzi,
  and David~J. Fleet.
\newblock Video diffusion models, 2022.

\bibitem{tifa}
Yushi Hu, Benlin Liu, Jungo Kasai, Yizhong Wang, Mari Ostendorf, Ranjay
  Krishna, and Noah~A Smith.
\newblock Tifa: Accurate and interpretable text-to-image faithfulness
  evaluation with question answering.
\newblock {\em arXiv preprint arXiv:2303.11897}, 2023.

\bibitem{kay2017kinetics}
Will Kay, Joao Carreira, Karen Simonyan, Brian Zhang, Chloe Hillier, Sudheendra
  Vijayanarasimhan, Fabio Viola, Tim Green, Trevor Back, Paul Natsev, et~al.
\newblock The kinetics human action video dataset.
\newblock {\em arXiv preprint arXiv:1705.06950}, 2017.

\bibitem{kendall1948rank}
Maurice~George Kendall.
\newblock Rank correlation methods.
\newblock 1948.

\bibitem{vae}
Diederik~P Kingma and Max Welling.
\newblock Auto-encoding variational bayes.
\newblock {\em arXiv preprint arXiv:1312.6114}, 2013.

\bibitem{klakow2002testing}
Dietrich Klakow and Jochen Peters.
\newblock Testing the correlation of word error rate and perplexity.
\newblock {\em Speech Communication}, 38(1-2):19--28, 2002.

\bibitem{blind_temporal}
Wei-Sheng Lai, Jia-Bin Huang, Oliver Wang, Eli Shechtman, Ersin Yumer, and
  Ming-Hsuan Yang.
\newblock Learning blind video temporal consistency.
\newblock In {\em Proceedings of the European conference on computer vision
  (ECCV)}, pages 170--185, 2018.

\bibitem{dvp}
Chenyang Lei, Yazhou Xing, and Qifeng Chen.
\newblock Blind video temporal consistency via deep video prior.
\newblock In {\em Advances in Neural Information Processing Systems}, 2020.

\bibitem{seedbench}
Bohao Li, Rui Wang, Guangzhi Wang, Yuying Ge, Yixiao Ge, and Ying Shan.
\newblock Seed-bench: Benchmarking multimodal llms with generative
  comprehension.
\newblock Jul 2023.

\bibitem{li2021unified}
Dingquan Li, Tingting Jiang, and Ming Jiang.
\newblock Unified quality assessment of in-the-wild videos with mixed datasets
  training.
\newblock {\em International Journal of Computer Vision}, 129:1238--1257, 2021.

\bibitem{li2023blip}
Junnan Li, Dongxu Li, Silvio Savarese, and Steven Hoi.
\newblock Blip-2: Bootstrapping language-image pre-training with frozen image
  encoders and large language models.
\newblock {\em arXiv preprint arXiv:2301.12597}, 2023.

\bibitem{wordnet}
George~A Miller.
\newblock Wordnet: a lexical database for english.
\newblock {\em Communications of the ACM}, 38(11):39--41, 1995.

\bibitem{morris2004and}
Andrew~Cameron Morris, Viktoria Maier, and Phil Green.
\newblock From wer and ril to mer and wil: improved evaluation measures for
  connected speech recognition.
\newblock In {\em Eighth International Conference on Spoken Language
  Processing}, 2004.

\bibitem{gpt4}
OpenAI.
\newblock Gpt-4 technical report, 2023.

\bibitem{paddleocr}
PaddlePaddle.
\newblock Paddleocr.
\newblock \url{https://github.com/PaddlePaddle/PaddleOCR}, 2013.

\bibitem{bleu}
Kishore Papineni, Salim Roukos, Todd Ward, and Wei-Jing Zhu.
\newblock {B}leu: a method for automatic evaluation of machine translation.
\newblock In {\em Proceedings of the 40th Annual Meeting of the Association for
  Computational Linguistics}, pages 311--318, Philadelphia, Pennsylvania, USA,
  July 2002. Association for Computational Linguistics.

\bibitem{sdxl}
Dustin Podell, Zion English, Kyle Lacey, Andreas Blattmann, Tim Dockhorn, Jonas
  M{\"u}ller, Joe Penna, and Robin Rombach.
\newblock Sdxl: improving latent diffusion models for high-resolution image
  synthesis.
\newblock {\em arXiv preprint arXiv:2307.01952}, 2023.

\bibitem{fatezero}
Chenyang Qi, Xiaodong Cun, Yong Zhang, Chenyang Lei, Xintao Wang, Ying Shan,
  and Qifeng Chen.
\newblock Fatezero: Fusing attentions for zero-shot text-based video editing.
\newblock {\em arXiv:2303.09535}, 2023.

\bibitem{clip}
Alec Radford, Jong~Wook Kim, Chris Hallacy, Aditya Ramesh, Gabriel Goh,
  Sandhini Agarwal, Girish Sastry, Amanda Askell, Pamela Mishkin, Jack Clark,
  et~al.
\newblock Learning transferable visual models from natural language
  supervision.
\newblock In {\em International conference on machine learning}, pages
  8748--8763. PMLR, 2021.

\bibitem{ldm}
Robin Rombach, Andreas Blattmann, Dominik Lorenz, Patrick Esser, and Björn
  Ommer.
\newblock High-resolution image synthesis with latent diffusion models, 2021.

\bibitem{imagen}
Chitwan Saharia, William Chan, Saurabh Saxena, Lala Li, Jay Whang, Emily~L
  Denton, Kamyar Ghasemipour, Raphael Gontijo~Lopes, Burcu Karagol~Ayan, Tim
  Salimans, et~al.
\newblock Photorealistic text-to-image diffusion models with deep language
  understanding.
\newblock {\em Advances in Neural Information Processing Systems},
  35:36479--36494, 2022.

\bibitem{is}
Tim Salimans, Ian Goodfellow, Wojciech Zaremba, Vicki Cheung, Alec Radford, and
  Xi Chen.
\newblock Improved techniques for training gans.
\newblock {\em Advances in neural information processing systems}, 29, 2016.

\bibitem{fid}
Maximilian Seitzer.
\newblock {pytorch-fid: FID Score for PyTorch}.
\newblock \url{https://github.com/mseitzer/pytorch-fid}, August 2020.
\newblock Version 0.3.0.

\bibitem{deepface}
Sefik~Ilkin Serengil and Alper Ozpinar.
\newblock Hyperextended lightface: A facial attribute analysis framework.
\newblock In {\em 2021 International Conference on Engineering and Emerging
  Technologies (ICEET)}, pages 1--4. IEEE, 2021.

\bibitem{shao2022overcoming}
Chenze Shao and Yang Feng.
\newblock Overcoming catastrophic forgetting beyond continual learning:
  Balanced training for neural machine translation.
\newblock {\em arXiv preprint arXiv:2203.03910}, 2022.

\bibitem{makeavideo}
Uriel Singer, Adam Polyak, Thomas Hayes, Xi Yin, Jie An, Songyang Zhang, Qiyuan
  Hu, Harry Yang, Oron Ashual, Oran Gafni, et~al.
\newblock Make-a-video: Text-to-video generation without text-video data.
\newblock {\em arXiv preprint arXiv:2209.14792}, 2022.

\bibitem{stylegan-v}
Ivan Skorokhodov, Sergey Tulyakov, and Mohamed Elhoseiny.
\newblock Stylegan-v: A continuous video generator with the price, image
  quality and perks of stylegan2.
\newblock In {\em Proceedings of the IEEE/CVF Conference on Computer Vision and
  Pattern Recognition}, pages 3626--3636, 2022.

\bibitem{sun2019icdar}
Yipeng Sun, Zihan Ni, Chee-Kheng Chng, Yuliang Liu, Canjie Luo, Chun~Chet Ng,
  Junyu Han, Errui Ding, Jingtuo Liu, Dimosthenis Karatzas, et~al.
\newblock Icdar 2019 competition on large-scale street view text with partial
  labeling-rrc-lsvt.
\newblock In {\em 2019 International Conference on Document Analysis and
  Recognition (ICDAR)}, pages 1557--1562. IEEE, 2019.

\bibitem{googlenet}
Christian Szegedy, Wei Liu, Yangqing Jia, Pierre Sermanet, Scott Reed, Dragomir
  Anguelov, Dumitru Erhan, Vincent Vanhoucke, and Andrew Rabinovich.
\newblock Going deeper with convolutions.
\newblock In {\em Proceedings of the IEEE conference on computer vision and
  pattern recognition}, pages 1--9, 2015.

\bibitem{raft}
Zachary Teed and Jia Deng.
\newblock Raft: Recurrent all-pairs field transforms for optical flow.
\newblock In {\em Computer Vision--ECCV 2020: 16th European Conference,
  Glasgow, UK, August 23--28, 2020, Proceedings, Part II 16}, pages 402--419.
  Springer, 2020.

\bibitem{llama}
Hugo Touvron, Thibaut Lavril, Gautier Izacard, Xavier Martinet, Marie-Anne
  Lachaux, Timoth{\'e}e Lacroix, Baptiste Rozi{\`e}re, Naman Goyal, Eric
  Hambro, Faisal Azhar, et~al.
\newblock Llama: Open and efficient foundation language models.
\newblock {\em arXiv preprint arXiv:2302.13971}, 2023.

\bibitem{llama2}
Hugo Touvron, Louis Martin, Kevin Stone, Peter Albert, Amjad Almahairi, Yasmine
  Babaei, Nikolay Bashlykov, Soumya Batra, Prajjwal Bhargava, Shruti Bhosale,
  et~al.
\newblock Llama 2: Open foundation and fine-tuned chat models.
\newblock {\em arXiv preprint arXiv:2307.09288}, 2023.

\bibitem{fvd}
Thomas Unterthiner, Sjoerd Van~Steenkiste, Karol Kurach, Raphael Marinier,
  Marcin Michalski, and Sylvain Gelly.
\newblock Towards accurate generative models of video: A new metric \&
  challenges.
\newblock {\em arXiv preprint arXiv:1812.01717}, 2018.

\bibitem{modelscope}
Jiuniu Wang, Hangjie Yuan, Dayou Chen, Yingya Zhang, Xiang Wang, and Shiwei
  Zhang.
\newblock Modelscope text-to-video technical report.
\newblock {\em arXiv preprint arXiv:2308.06571}, 2023.

\bibitem{wang2023videomaev2}
Limin Wang, Bingkun Huang, Zhiyu Zhao, Zhan Tong, Yinan He, Yi Wang, Yali Wang,
  and Yu Qiao.
\newblock Videomae v2: Scaling video masked autoencoders with dual masking,
  2023.

\bibitem{dover}
Haoning Wu, Erli Zhang, Liang Liao, Chaofeng Chen, Jingwen~Hou Hou, Annan Wang,
  Wenxiu~Sun Sun, Qiong Yan, and Weisi Lin.
\newblock Exploring video quality assessment on user generated contents from
  aesthetic and technical perspectives.
\newblock In {\em International Conference on Computer Vision (ICCV)}, 2023.

\bibitem{vdmsurvey}
Zhen Xing, Qijun Feng, Haoran Chen, Qi Dai, Han Hu, Hang Xu, Zuxuan Wu, and
  Yu-Gang Jiang.
\newblock A survey on video diffusion models.
\newblock {\em arXiv preprint arXiv:2310.10647}, 2023.

\bibitem{tooleval}
Qiantong Xu, Fenglu Hong, Bo Li, Changran Hu, Zhengyu Chen, and Jian Zhang.
\newblock On the tool manipulation capability of open-source large language
  models, 2023.

\bibitem{ye2023mplug}
Qinghao Ye, Haiyang Xu, Guohai Xu, Jiabo Ye, Ming Yan, Yiyang Zhou, Junyang
  Wang, Anwen Hu, Pengcheng Shi, Yaya Shi, et~al.
\newblock mplug-owl: Modularization empowers large language models with
  multimodality.
\newblock {\em arXiv preprint arXiv:2304.14178}, 2023.

\bibitem{digan}
Sihyun Yu, Jihoon Tack, Sangwoo Mo, Hyunsu Kim, Junho Kim, Jung-Woo Ha, and
  Jinwoo Shin.
\newblock Generating videos with dynamics-aware implicit generative adversarial
  networks.
\newblock In {\em International Conference on Learning Representations}, 2022.

\bibitem{zar2005spearman}
Jerrold~H Zar.
\newblock Spearman rank correlation.
\newblock {\em Encyclopedia of Biostatistics}, 7, 2005.

\bibitem{zhang2023show}
David~Junhao Zhang, Jay~Zhangjie Wu, Jia-Wei Liu, Rui Zhao, Lingmin Ran, Yuchao
  Gu, Difei Gao, and Mike~Zheng Shou.
\newblock Show-1: Marrying pixel and latent diffusion models for text-to-video
  generation.
\newblock {\em arXiv preprint arXiv:2309.15818}, 2023.

\bibitem{sadtalker}
Wenxuan Zhang, Xiaodong Cun, Xuan Wang, Yong Zhang, Xi Shen, Yu Guo, Ying Shan,
  and Fei Wang.
\newblock Sadtalker: Learning realistic 3d motion coefficients for stylized
  audio-driven single image talking face animation, 2022.

\bibitem{zhao2023survey}
Wayne~Xin Zhao, Kun Zhou, Junyi Li, Tianyi Tang, Xiaolei Wang, Yupeng Hou,
  Yingqian Min, Beichen Zhang, Junjie Zhang, Zican Dong, et~al.
\newblock A survey of large language models.
\newblock {\em arXiv preprint arXiv:2303.18223}, 2023.

\bibitem{zheng2023judging}
Lianmin Zheng, Wei-Lin Chiang, Ying Sheng, Siyuan Zhuang, Zhanghao Wu, Yonghao
  Zhuang, Zi Lin, Zhuohan Li, Dacheng Li, Eric.~P Xing, Hao Zhang, Joseph~E.
  Gonzalez, and Ion Stoica.
\newblock Judging llm-as-a-judge with mt-bench and chatbot arena, 2023.

\bibitem{magicvideo}
Daquan Zhou, Weimin Wang, Hanshu Yan, Weiwei Lv, Yizhe Zhu, and Jiashi Feng.
\newblock Magicvideo: Efficient video generation with latent diffusion models.
\newblock {\em arXiv preprint arXiv:2211.11018}, 2022.

\bibitem{gu2023xiezhi}
Gu Zhouhong, Zhu Xiaoxuan, Ye Haoning, Zhang Lin, Wang Jianchen, Jiang Sihang,
  Xiong Zhuozhi, Li Zihan, He Qianyu, Xu Rui, Huang Wenhao, Zheng Weiguo, Feng
  Hongwei, and Xiao Yanghua.
\newblock Xiezhi: An ever-updating benchmark for holistic domain knowledge
  evaluation.
\newblock {\em arXiv:2304.11679}, 2023.

\end{thebibliography}
}

\ifarxiv 
\clearpage 
\appendix

\begin{appendices}

\addcontentsline{toc}{section}{Appendix} 
\section*{Appendix Contents}
\startcontents[appendices]
\printcontents[appendices]{l}{1}{\setcounter{tocdepth}{2}}



\section{Detailed Analysis of Real-World User Data}
\label{sec:Real-World Data Analysis}


\begin{figure*}[h!]
    \centering
    \begin{subfigure}[b]{0.28\textwidth}
        \includegraphics[width=\textwidth]{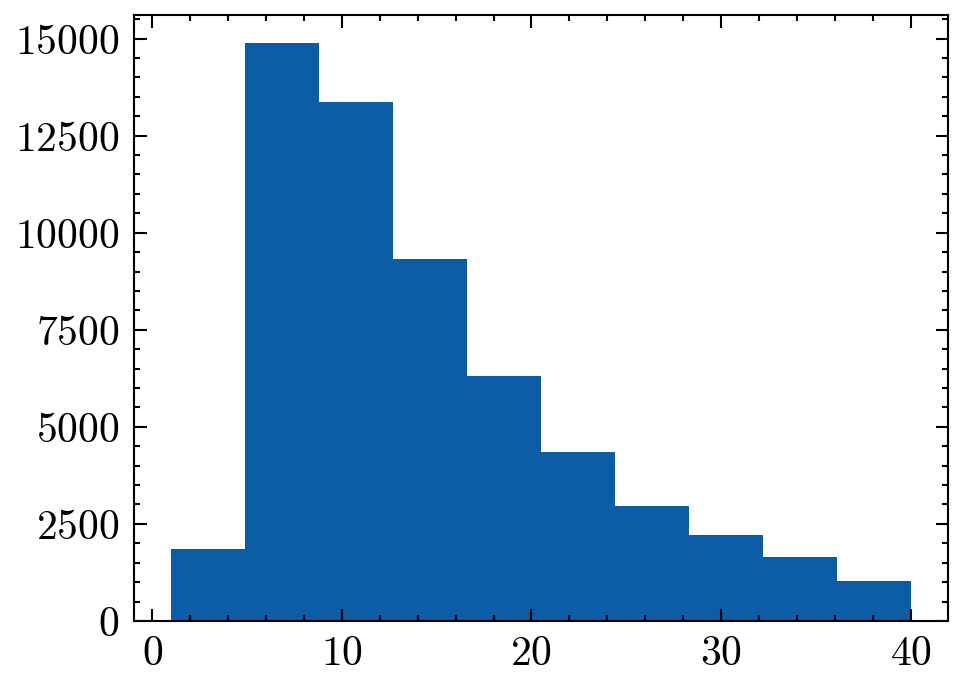}
        \caption{Prompt length distribution.}
        \label{fig:word_count}
    \end{subfigure}
    \hfill
    \begin{subfigure}[b]{0.42\textwidth}
        \includegraphics[width=\textwidth]{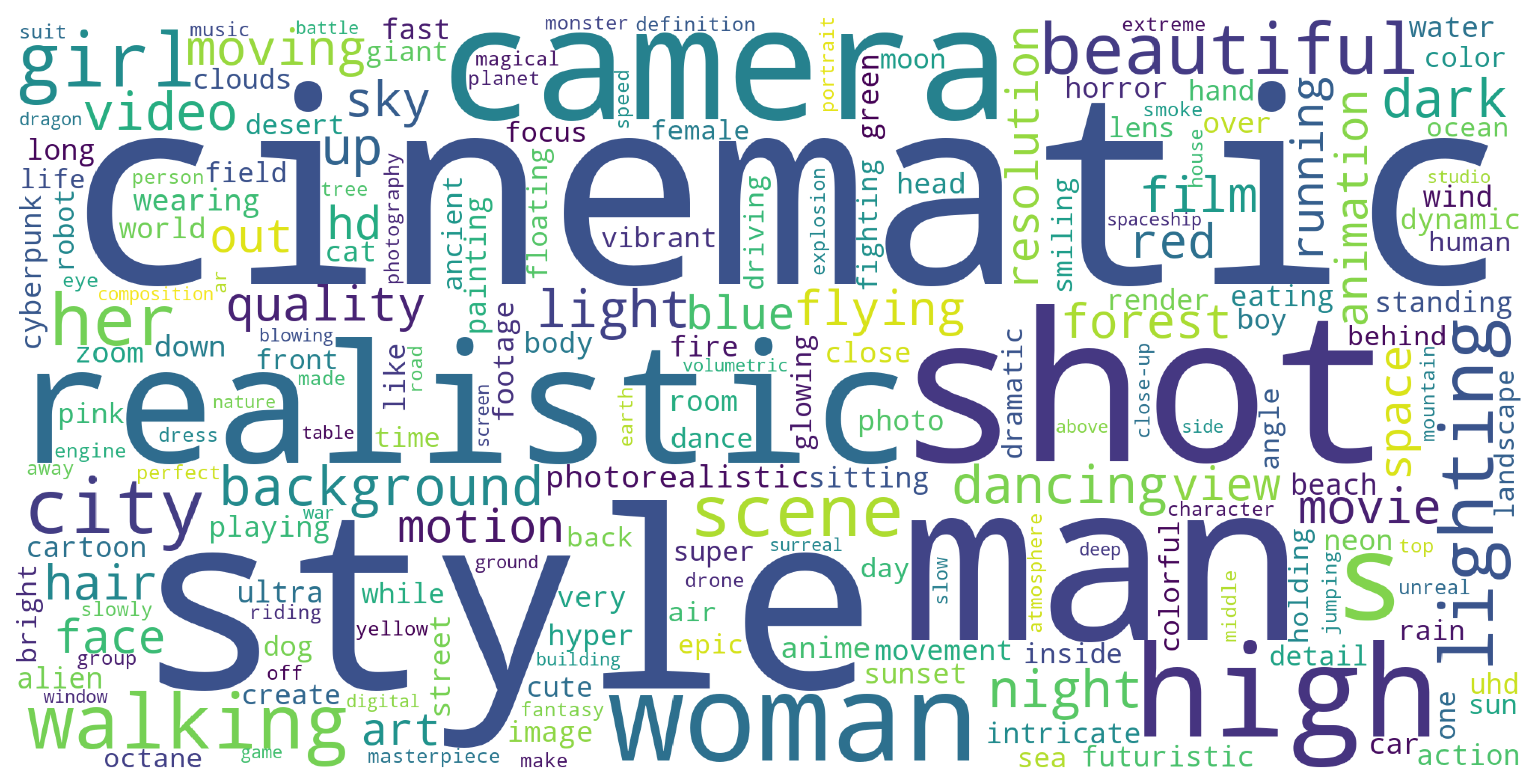}
        \caption{All words cloud}
        \label{fig:word_cloud}
    \end{subfigure}
    \hfill
    \begin{subfigure}[b]{0.28\textwidth}
        \includegraphics[width=\textwidth]{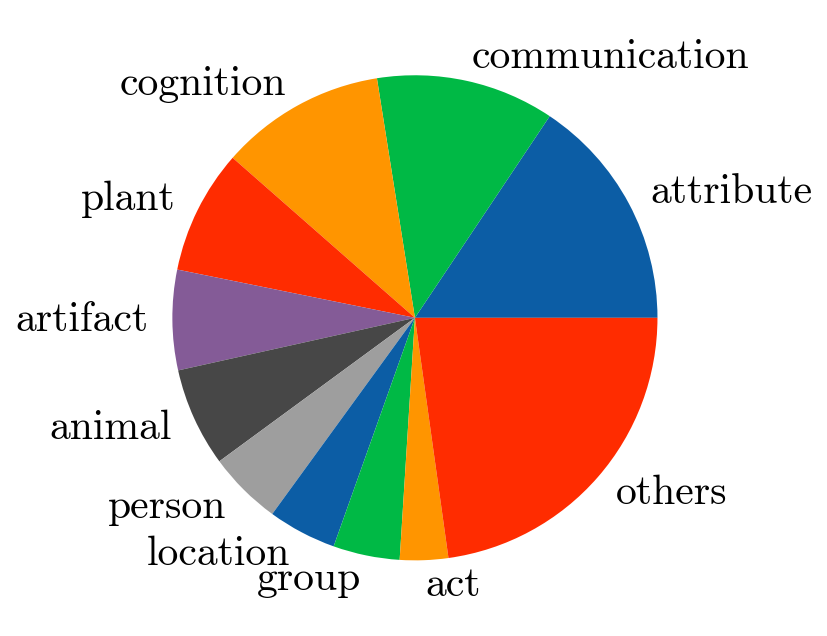}
        \caption{types of noun}
        \label{fig:word_types}
    \end{subfigure}
    \vspace{-1em}
    \caption{The analysis of the real-world prompts from PikaLab Server~\cite{pika}.}
    \vspace{-1em}
    \label{fig:analysis}
\end{figure*}

In this section, we present a detailed analysis of the real-world user data collected from text-to-video (T2V) generation discord users, including the FullJourney~\cite{fulljourney} and PikaLab~\cite{pika}. We provide insights into the distribution of prompt lengths, important words, and meta classes.

\subsection{Prompt Length Distribution}

Fig.~\ref{fig:analysis}~\subref{fig:word_count} shows the distribution of prompt lengths in the real-world user data. We find that 90\% of the prompts contain words in the range of $[3, 40]$. This observation helps us determine the appropriate length for the prompts in our benchmark.

\subsection{Important Words in Prompts}

Fig.~\ref{fig:analysis}~\subref{fig:word_cloud} presents a word cloud of all words in the real-world user data. From this word cloud, we can observe the most frequent words in the prompts and gain insights into the key concepts that users request in T2V generation.

\subsection{Meta Classes in Prompts}

Fig.~\ref{fig:analysis}~\subref{fig:word_types} shows the distribution of noun types in the real-world user data. We use WordNet~\cite{wordnet} to identify the meta classes. Excluding communication, attribute, and cognition words, we find that artifacts (human-made objects), humans, animals, and locations (landscapes) play important roles in the prompts. We also include the most important word \texttt{style} from Fig.~\ref{fig:analysis}~\subref{fig:word_cloud} in the meta classes.

Based on this analysis, we divide the T2V generation into four meta-subject classes: \texttt{human}, \texttt{animal}, \texttt{object}, and \texttt{landscape}. This classification helps us create a diverse and representative benchmark for evaluating T2V models.

\begin{table*}[t]
\centering
\begin{tabular}{c|c|c|c|c|c|c|c|c}
\toprule
 Metrics & Gen2 & ModelScope & Pika & Floor33 & ZeroScope & VideoCrafter & Show-1 & Hotshot \\ \hline \hline
VQA$_A$ $\uparrow$  & 59.44 & 40.06 & 59.09 & 58.7 & 34.02 & 66.18 & 23.19 & 71.54 \\ \hline
VQA$_T$ $\uparrow$ & 76.51 & 32.93 & 64.96 & 52.64 & 39.94 & 58.93 & 44.24 & 50.52 \\ \hline
IS $\uparrow$  & 14.53 & 17.64 & 14.81 & 17.01 & 14.48 & 16.43 & 17.65 & 17.29 \\ \hline
CLIP-Temp $\uparrow$  & 99.94 & 99.74 & 99.97 & 99.6 & 99.84 & 99.78 & 99.77 & 99.74 \\ \hline
Warping Error  $\downarrow$  & 0.0008 & 0.0162 & 0.0006 & 0.0413 & 0.0193 & 0.0295 & 0.0067 & 0.0091 \\ \hline
Face Consistency $\uparrow$ & 99.06 & 98.94 & 99.62 & 99.08 & 99.33 & 99.48 & 99.32 & 99.48 \\ \hline
Action-Score  $\uparrow$ & 62.53 & 72.12 & 71.81 & 71.66 & 67.56 & 68.06 & 81.56 & 66.8 \\ \hline
Motion AC-Score $\uparrow$ & 44.0 & 42.0 & 44.0 & 74.0 & 50.0 & 50.0 & 50.0 & 56.0 \\ \hline
Flow-Score $\rightarrow$ & 0.7 & 6.99 & 0.5 & 9.26 & 4.5 & 5.44 & 2.07 & 5.06 \\ \hline
CLIP-Score $\uparrow$ & 20.53 & 20.36 & 20.46 & 21.02 & 20.2 & 21.33 & 20.66 & 20.33 \\ \hline
BLIP-BLUE $\uparrow$ & 22.24 & 22.54 & 21.14 & 22.73 & 21.2 & 22.17 & 23.24 & 23.59 \\ \hline
SD-Score $\uparrow$ & 68.58 & 67.93 & 68.57 & 68.7 & 67.79 & 68.73 & 68.42 & 67.65 \\ \hline
Detection-Score $\uparrow$ & 64.05 & 50.01 & 58.99 & 52.44 & 53.94 & 67.67 & 58.63 & 45.7 \\ \hline
Color-Score $\uparrow$ & 37.56 & 38.72 & 34.35 & 41.85 & 39.25 & 45.11 & 48.55 & 42.39 \\ \hline
Count-Score $\uparrow$ & 53.31 & 44.18 & 51.46 & 58.33 & 41.01 & 58.11 & 44.31 & 49.5 \\ \hline
OCR-Score  $\downarrow$ & 75.0 & 71.32 & 84.31 & 87.48 & 82.58 & 88.04 & 58.97 & 63.66 \\ \hline
Celebrity ID Score $\downarrow$   & 41.25 & 44.56 & 45.21 & 40.07 & 46.93 & 40.18 & 37.93 & 38.58 \\ \bottomrule
\end{tabular}
\caption{
Raw results of 17 introduced metrics among the aspects of video quality, text-video alignment, motion quality, and temporal consistency. All metrics are expressed as percentages, except for Warping Error and Flow-Score.
}
\label{tab:results}
\end{table*}

\section{Quantitative Results}
\label{sec:Quantative Results}

In this part, we present the quantitative results of our evaluation benchmark. We have conducted experiments on various state-of-the-art video generative models and assessed their performance using 17 objective metrics.  We provide the raw results of every metric for each model and the correlations between metrics and human labels. The results are illustrated in two tables. The first table (Table \ref{tab:results}) shows the raw results of every metric for each model. The second table (Table \ref{tab:correlations_whole}) displays the correlations between metrics and human labels.

\subsection{Raw Results of Every Metric for Every Model}

Table \ref{tab:results} shows the raw results of all 17 introduced metrics for each of the evaluated models. All metrics are expressed as percentages, except for Warping Error and Flow-Score. The table is organized as follows:

\begin{itemize}
    \item The first column lists the metrics used for evaluation.
    \item The following columns display the raw results for each model, including ModelScope~\cite{modelscope}, Floor33 Pictures~\cite{floor33}, and ZeroScope~\cite{zeroscope}, Show-1~\cite{zhang2023show}, Hotshot-XL~\cite{Hotshot-XL}, VideoCrafter1~\cite{chen2023videocrafter1}, Gen2~\cite{gen1}, and PikaLab~\cite{pika}.
    \item Arrows next to the metric names indicate whether higher ($\uparrow$) or lower ($\downarrow$) values are better for that particular metric. For Flow-Score, the arrow is replaced with a rightwards arrow ($\rightarrow$) as it is a neutral metric.
\end{itemize}
\begin{table}[t]
\centering

\begin{tabular}{cl|cc}
\toprule
 Aspects & Methods & Spearman & Kendall \\
\hline
\multirow{5}{*}{\shortstack{Visual\\Quality}}   & VQA$_A$ & 47.8  &  35.5    \\
                                  & VQA$_T$ & 53.6  &  39.1    \\
                                  & IS & 9.9 & 4.3 \\
                                  & Avg.    & 54.9  &  40.9    \\
                                  & \textbf{Ours}    & \textbf{55.4}  &  \textbf{41.1}    \\ \hline\hline

\multirow{5}{*}{\shortstack{Motion\\Amplitude}}       
                                  & Action-Score & -14.9  & -10.4    \\
                                  & Motion AC & -22.1  & -16.4    \\
                                  & Flow-Score & -43.3  & -30.1    \\
                                  & Avg.     & -38.2  & -27.7    \\
                                  &\textbf{Ours}       & \textbf{45.0}  &  \textbf{32.4}    \\ \hline\hline

\multirow{5}{*}{\shortstack{Temporal\\Consistency}} & CLIP-Temp & 49.7  & 35.7    \\
                          & Warping Error & 69.0  & 51.7   \\
                          & Face Consistency & 25.8 & 17.8 \\
                          & Avg.     &   54.4  & 38.9    \\
                          & \textbf{Ours}       & \textbf{56.7}  &  \textbf{41.5}  \\ \hline\hline

\multirow{9}{*}{\shortstack{TV\\Alignment}} 
                        & CLIP-Score   & 6.3  & 4.3  \\
                          &  BLIP-BLEU   & 26.7  & 19.0  \\
                          & SD-Score & -2.8  & -2.3  \\
                          & Detection-Score & 11.9 & 9.4 \\
                          & Color-Score & -5.5 & -3.9 \\
                          & Count-Score & 28.9 & 22.2 \\
                          & OCR-Score & -8.3 & -6.7 \\
                          & Celebrity ID Score & -26.0 & -19.8 \\
                          & Avg.         & 31.9 & \textbf{22.7} \\
                          & \textbf{Ours}  & \textbf{32.3}  & 22.5  \\

\bottomrule
\end{tabular}
\vspace{-1em}
\caption{\textbf{\textit{Correlation Analysis.}}
Whole results of correlations between objective metrics and human judgment on T2V generations. We use Spearman's $\rho$ and Kendall's $\phi$ for correlation calculation.
}
\vspace{-1em}
\label{tab:correlations_whole}
\end{table}
\subsection{Correlations Between Metrics and Human Labels}

In addition to the raw results, Table \ref{tab:correlations_whole} presents the correlation analysis between objective metrics and human judgment on T2V generations. We use Spearman's $\rho$ and Kendall's $\phi$ for correlation calculation. The table is organized into four sections, representing the four aspects of the evaluation: visual quality, motion amplitude, temporal consistency, and text-video alignment. In each section, we compare various methods with our proposed evaluation method, which is highlighted in bold.

As can be seen from the table, our method consistently achieves higher correlation values compared to the average of other methods. This shows the effectiveness of our proposed evaluation method in aligning the objective metrics to users' opinions. For instance, in the visual quality aspect, our method obtains a Spearman's $\rho$ of 55.4 and a Kendall's $\phi$ of 41.1, which are both higher than the average values of 55.0 and 41.0, respectively. Similar improvements can be observed in other aspects as well.

In addition to the findings mentioned earlier, we can observe that some metrics show negative correlations with human judgment, such as Color-Score and OCR-Score in the TV Alignment aspect. This indicates that these metrics may not be reliable for evaluating the alignment between text and video content in generative models. On the other hand, metrics like Detection-Score and Count-Score exhibit relatively higher correlations with human judgment, suggesting their potential usefulness in evaluating T2V alignment.

Overall, the results in Table \ref{tab:correlations_whole} provide a comprehensive analysis of various objective metrics and their correlations with human judgment. These results can be valuable for researchers and practitioners in the field of T2V generation to select appropriate metrics for evaluating their models and to better understand the strengths and weaknesses of different evaluation methods.

\section{Qualitative Results}
\label{sec:Qualitative Results}

In this part, we present qualitative results of the evaluated T2V models for various aspects of video generation, taking into account the findings listed in the paper. The results are visualized in Fig. \ref{fig:Vid_Gen_Style} to Fig. \ref{fig:Vid_Gen_Task}. We discuss the performance of each model in terms of camera motion control, content generation, motion generation, style generation, and task-specific generation. 

\subsection{ Content Generation}
In Fig. \ref{fig:Vid_Gen_Content}, we present the qualitative results of T2V models and SDXL~\cite{sdxl} for four meta types of content generation: human, object, landscape, and animal. Finding \#5 shows that resolution does not correlate much with visual appeal, as demonstrated by Gen2~\cite{gen1} and Hotshot-XL~\cite{Hotshot-XL}, which have small resolutions but are both competitive in visual quality. Besides, we can also find that Gen2~\cite{gen1} and PikaLab~\cite{pika} are more distinguishable from SDXL~\cite{sdxl} in both video content and style compared with other methods.

\subsection{Motion Generation}
Fig. \ref{fig:Vid_Gen_Motion} displays the qualitative results of T2V models with respect to motion generation. According to Finding \#6, larger motion amplitude does not ensure user preference. In our study, most videos that users are fond of are those with slight movements, such as those generated by PikaLab~\cite{pika} and Gen2~\cite{gen1}.

\subsection{Style Generation}
The qualitative results of T2V models concerning style generation are shown in Fig. \ref{fig:Vid_Gen_Style}. We can see from the figure that most methods have the ability to generate videos with specific styles, which may be inherited from base models. However, various methods like  ZeroScope~\cite{zeroscope} and ModelScope~\cite{modelscope} are also struggling to generate high-quality and consistent styled content from prompts.

\subsection{Camera Motion Control}
Fig. \ref{fig:Vid_Gen_Camera_Motion} shows the qualitative results of T2V models in terms of prompts with camera motion controls. As indicated by Finding \#4, all methodss cannot perform camera motion control using text prompts, which indicates all T2V models lack the understanding of camera motion. 

\subsection{Task-Specific Generation}
Finally, Fig. \ref{fig:Vid_Gen_Task} presents the qualitative results of T2V models and SDXL~\cite{sdxl} in terms of different tasks,  \ie, face generation, object generation with color, object generation with count, text generation, and activity generation. Finding \#8 indicates that many models can sometimes generate completely wrong videos, with severe noises and distortions observed in baseline models like ZeroScope~\cite{zeroscope}, ModelScope~\cite{modelscope}, and Floor33 Pictures~\cite{floor33}. This could be viewed as a catastrophic forgetting problem, as many current T2V models are finetuned from base models like SD~\cite{ldm}.

In conclusion, the qualitative results presented in this appendix provide valuable insights into the strengths and weaknesses of different T2V models in various aspects of video generation. As stated in Finding \#10, all current models are not satisfactory enough, and T2V models still have significant room for improvement. Even the best model in our evaluation, Gen2~\cite{gen1}, has limitations like struggling with complex scenes, instruction following, and entity details. These results, along with our proposed evaluation framework and pipeline, enable a more exhaustive and reliable assessment of the performance of large video generation models.

\section{Additional Analysis and Explanations}
\noindent \subsection{Adequacy of 700 Prompts}
As an initial attempt, one important reason to use these prompts is that we find the metrics tend to reach a plateau with the sample increased as in Fig.~\ref{fig:clip}. Besides, concurrent benchmarks~(\eg, FETV~(619 prompts overall), VBench~(100 prompts per metric)) use a similar amount of prompts. 
From a practical view, T2V models' sampling speed is typically slow, a small but effective benchmark is crucial for fast evaluation of different methods. 

\noindent {\subsection{User study’s Demographics, Interface, and Adequacy} }
The interface is shown below. We use our internal AI-testing platform to find human raters. Each rater is asked to perform 100 tasks as pre-labeling, the annotator who has more than 90\% accuracy will be marked as qualified. Otherwise, we will consider another supplier. Then, the qualified annotators will label the whole benchmark.

\begin{figure}[h]
    \centering
    \vspace{-1em}
    \includegraphics[width=0.9\linewidth]{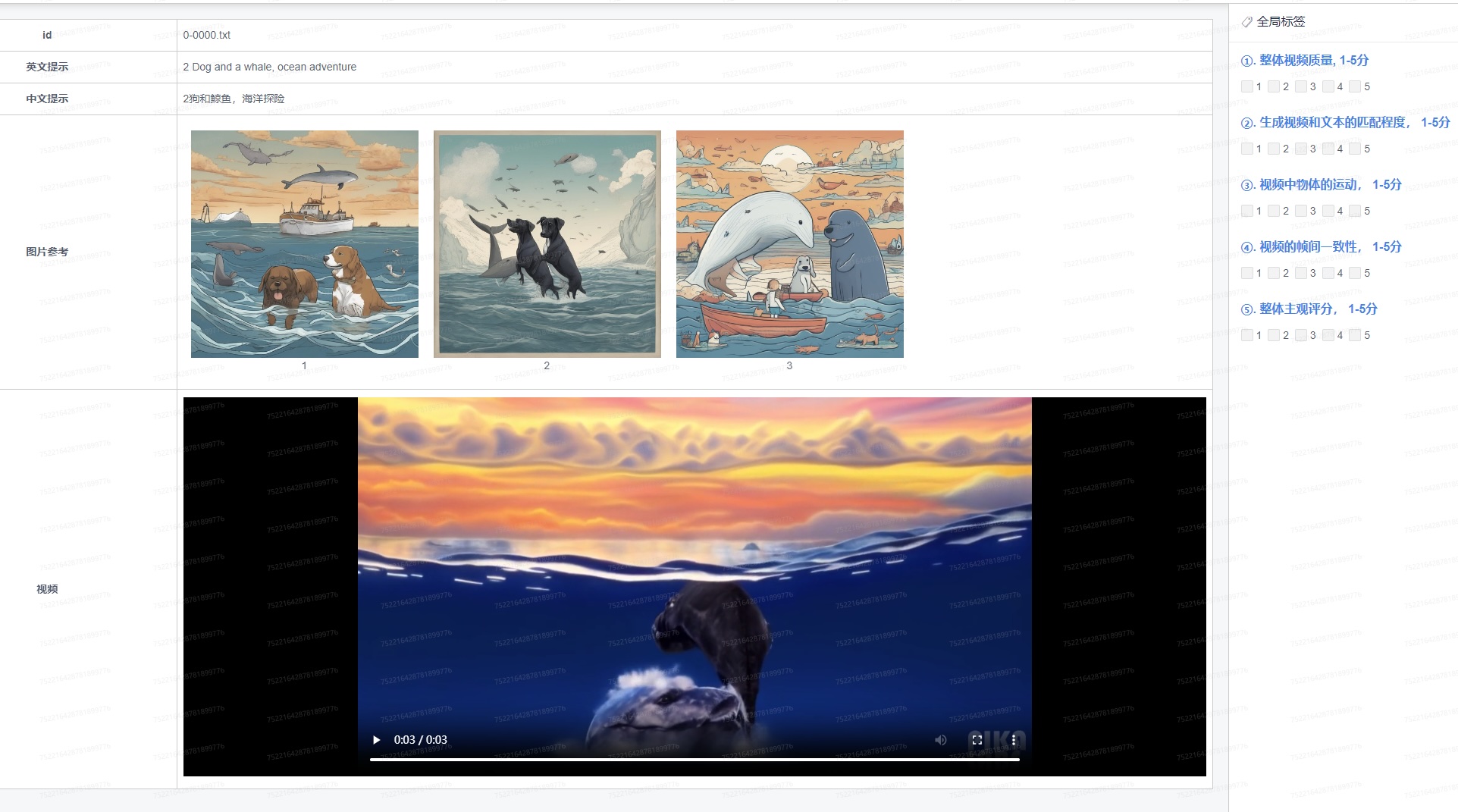}
    \vspace{-2em}
    \label{fig:user_study}
\end{figure}

\noindent \subsection{Crowdsourced Prompts \& Dataset Dynamics}
Our approach aims to capture a snapshot of current user expectations and model abilities. Besides, the benchmark is intended to be dynamic with periodic updates.

\noindent \subsection{Motion Amplitude Metrics Discrepancy}
It's mainly caused by user preferences on favoring subtle motions as stated by Finding \#6, \eg, Gen2 (always generate small motions) ranks 7th \textit{w.r.t.} Motion AC-Score in Tab. 4, but it ranks 1st \textit{w.r.t.} motion quality in user study, which resulted in Gen2 ranks 1st in Tab. 2. 

\noindent \subsection{Dependence on Pre-trained Models}
As our initial attempt~(also the whole community), pre-trained models provide a reference to find meaningful objective metrics. We will actively explore more straightforward metrics to avoid using pre-trained models, \eg, training an end-to-end evaluation model for each aspect using more user opinions.

\noindent \subsection{Costs of Evaluation} 
We agree that online evaluation is vital for model training. However, it is impossible to monitor T2V Models in the running process~(even using FVD) since each video sample requires more than 2 minutes for generation. Our method is designed for offline evaluation~(plays a similar role to previous FVD evaluation). The whole benchmark requires around 2 hours on an A100 GPU and at last 16 GB memory cost without any code optimization, which we think is more demanding than traditional methods such as FVD. However, FVD can only reflect one aspect of the T2V model and needs real video datasets as a reference.

\begin{figure}[tp]
    \centering
    \includegraphics[width=0.8\linewidth]{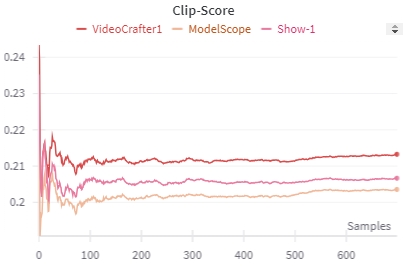}
    \vspace{-1em}
    \caption{The stability of CLIP-Score with different prompts.}
    \vspace{-2em}
    \label{fig:clip}
\end{figure}

\noindent \subsection{Benchmark Stability and Noise Variables}
We give some stability analysis. First, as in Fig.~\ref{fig:clip}, we find the objective scores are stable among different methods with the prompt increasing. 
Besides, we also try to introduce noise to the prompts, \ie, adding, removing, or swapping words/symbols in 100 randomly selected prompts. 
Notably, the changes in all scores in Tab. 2 are marginal among the methods, with most variations below 0.2 points. The ranks remain consistent across all models.

\noindent  \subsection{Warping Error}
The warping error assesses the discrepancy between the actual subsequent frame and its prediction, generated by warping the current frame using optical flow.
The larger warping error means each frame changes dramatically, which is typically unwanted for real-world video. It also widely used previous blind video consistency methods~\cite{blind_temporal} for temporal consistency metrics.

\noindent \subsection{Discrepancy in User Study and Evaluation}
This discrepancy arose from our endeavors to continuously update and enhance our prompt list. Our initial benchmark contains 512 prompts for user study, and we further expanded it to 700 prompts to make it more comprehensive and balanced. However, similar to Fig.~\ref{fig:clip}, there are no significant changes in our results after the prompt increment. Therefore, we use the same user study result to avoid wasting resources as the initial version.

\begin{figure*}[tp]
    \centering
    \includegraphics[width=\linewidth]{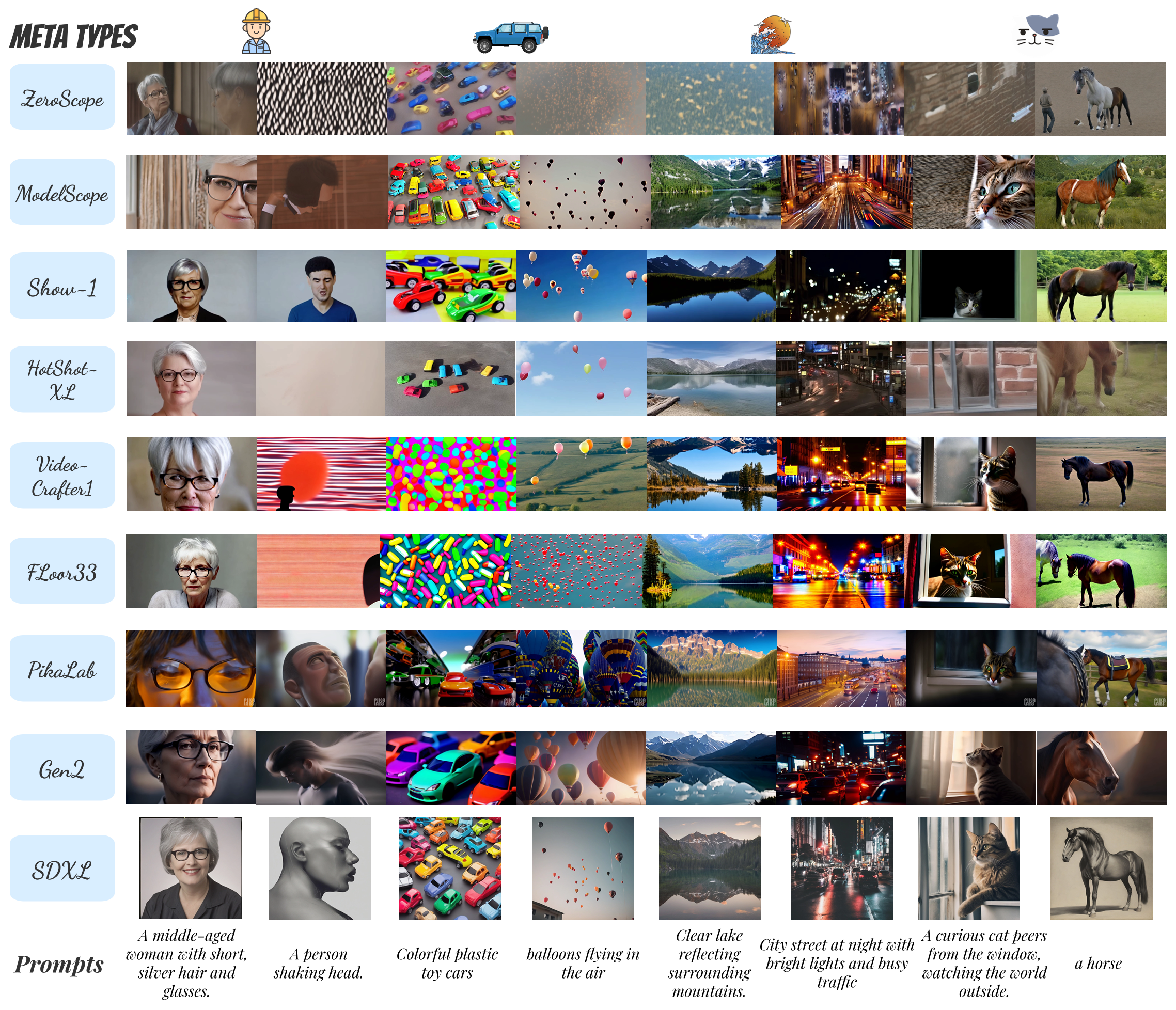}
    \vspace{-2em}
    \caption{Qualitative results of T2V models in terms of four meta types (i.e., human, object, landscape, and animal)}
    \label{fig:Vid_Gen_Content}
\end{figure*}

\begin{figure*}[tp]
    \centering
    \includegraphics[width=\linewidth]{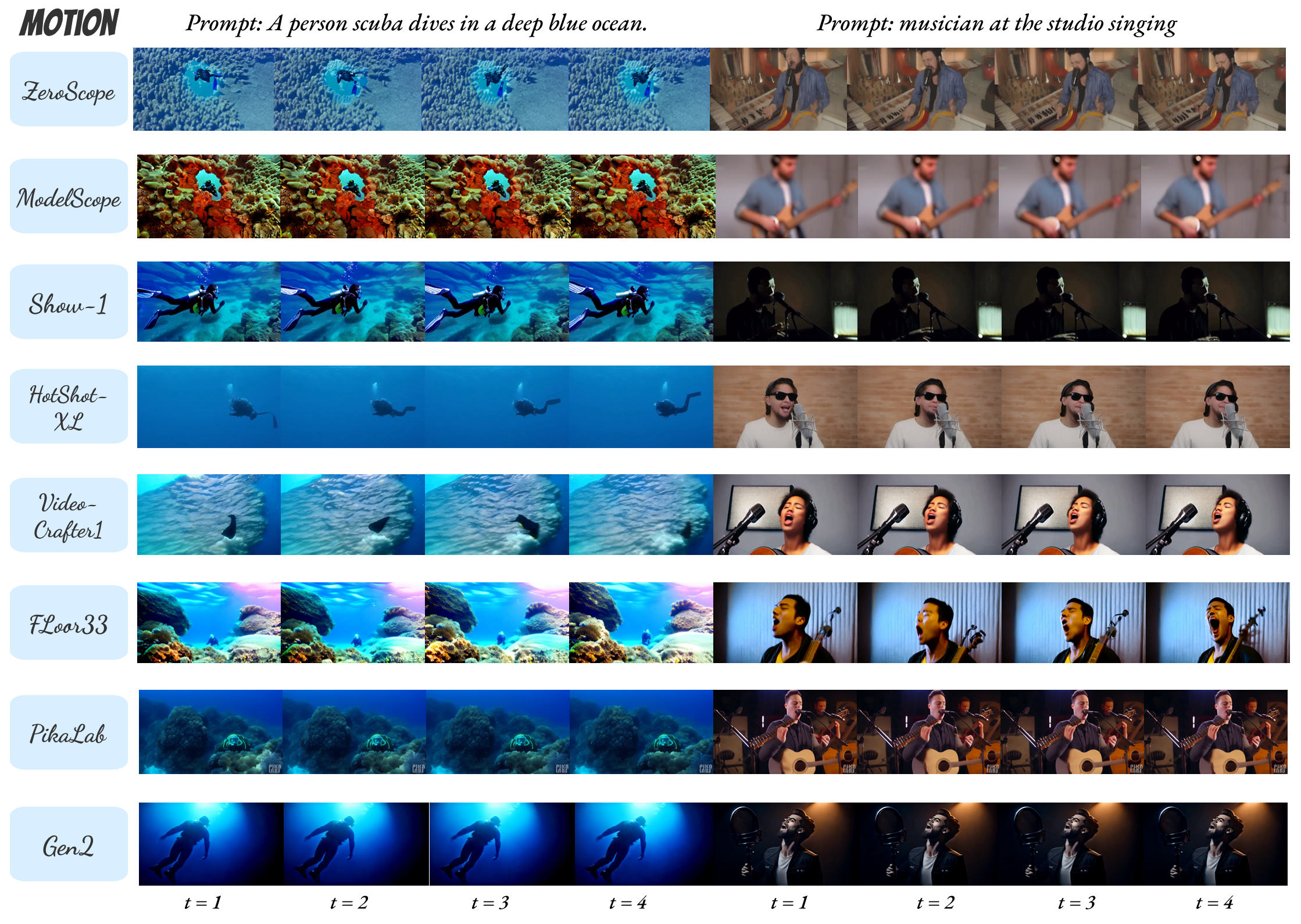}
    \vspace{-2em}
    \caption{Qualitative results of T2V models w.r.t. motion generation}
    \label{fig:Vid_Gen_Motion}
\end{figure*}

\begin{figure*}[tp]
    \centering
    \includegraphics[width=\linewidth]{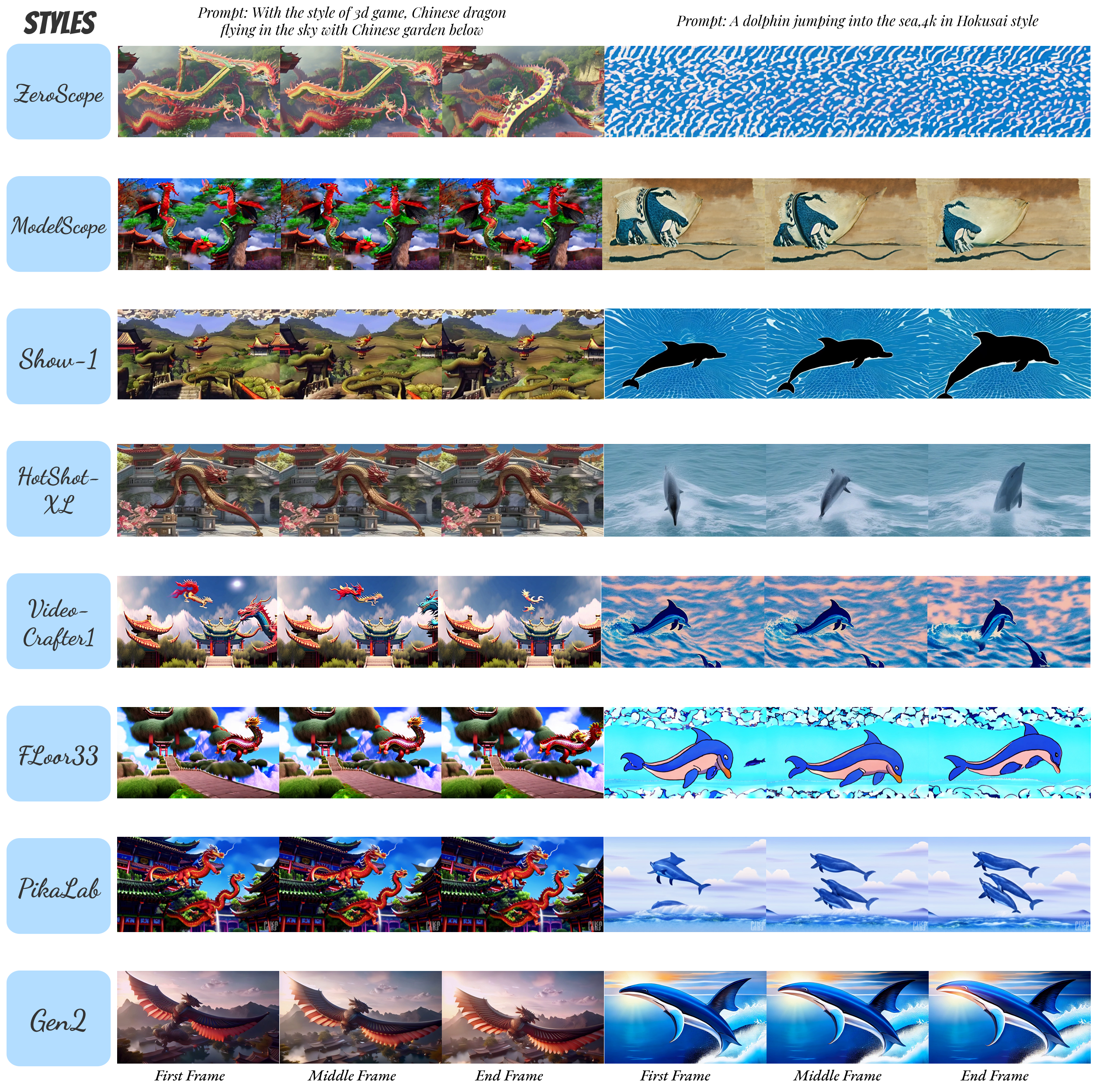}
    \vspace{-2em}
    \caption{Qualitative results of T2V models  w.r.t. style generation}
    \label{fig:Vid_Gen_Style}
\end{figure*}

\begin{figure*}[tp]
    \centering
    \includegraphics[width=\linewidth]{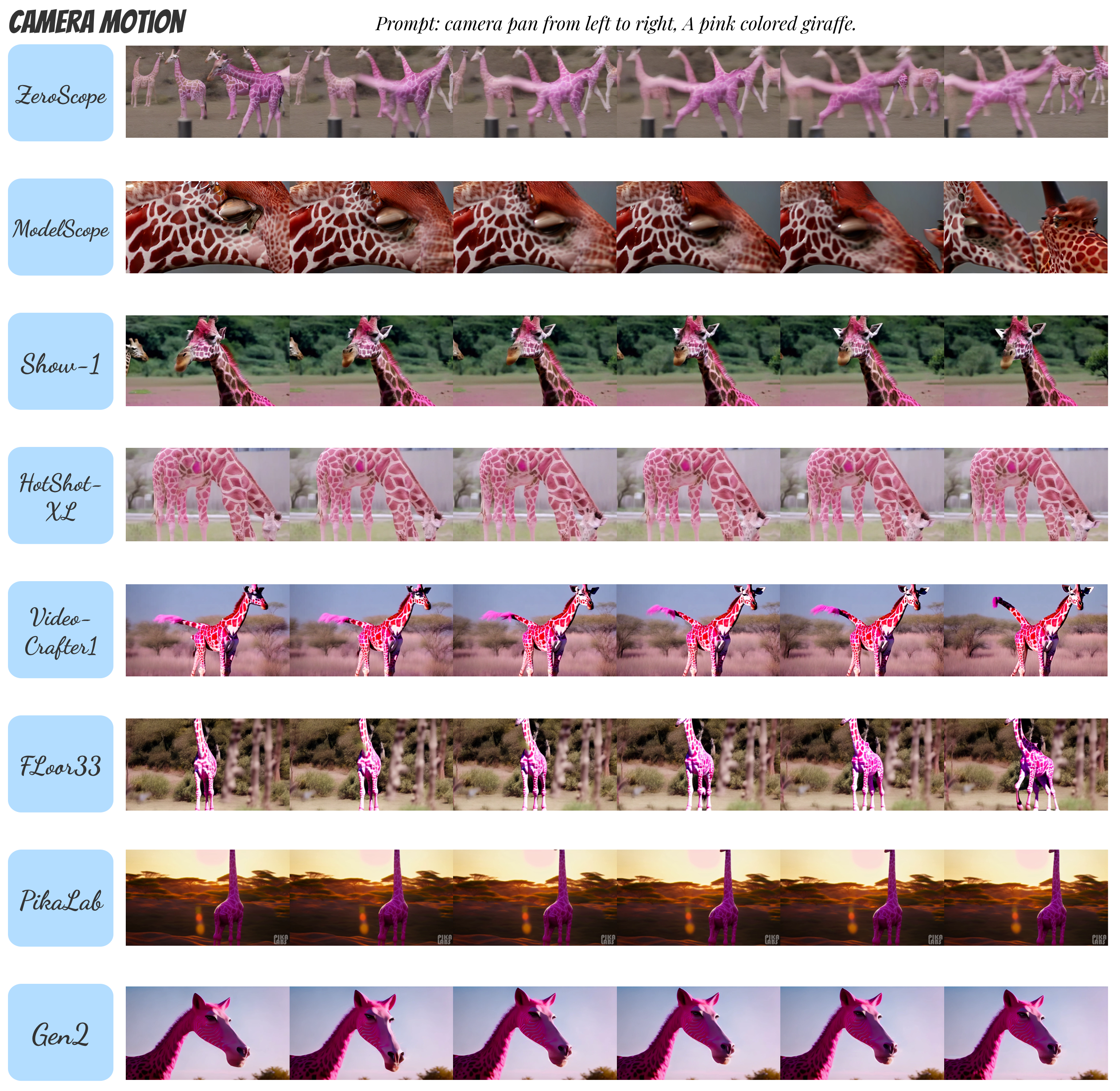}
    \vspace{-2em}
    \caption{Qualitative results of T2V models in terms of prompts wtih camera motion controls}
    \label{fig:Vid_Gen_Camera_Motion}
\end{figure*}

\begin{figure*}[tp]
    \centering
    \includegraphics[width=\linewidth]{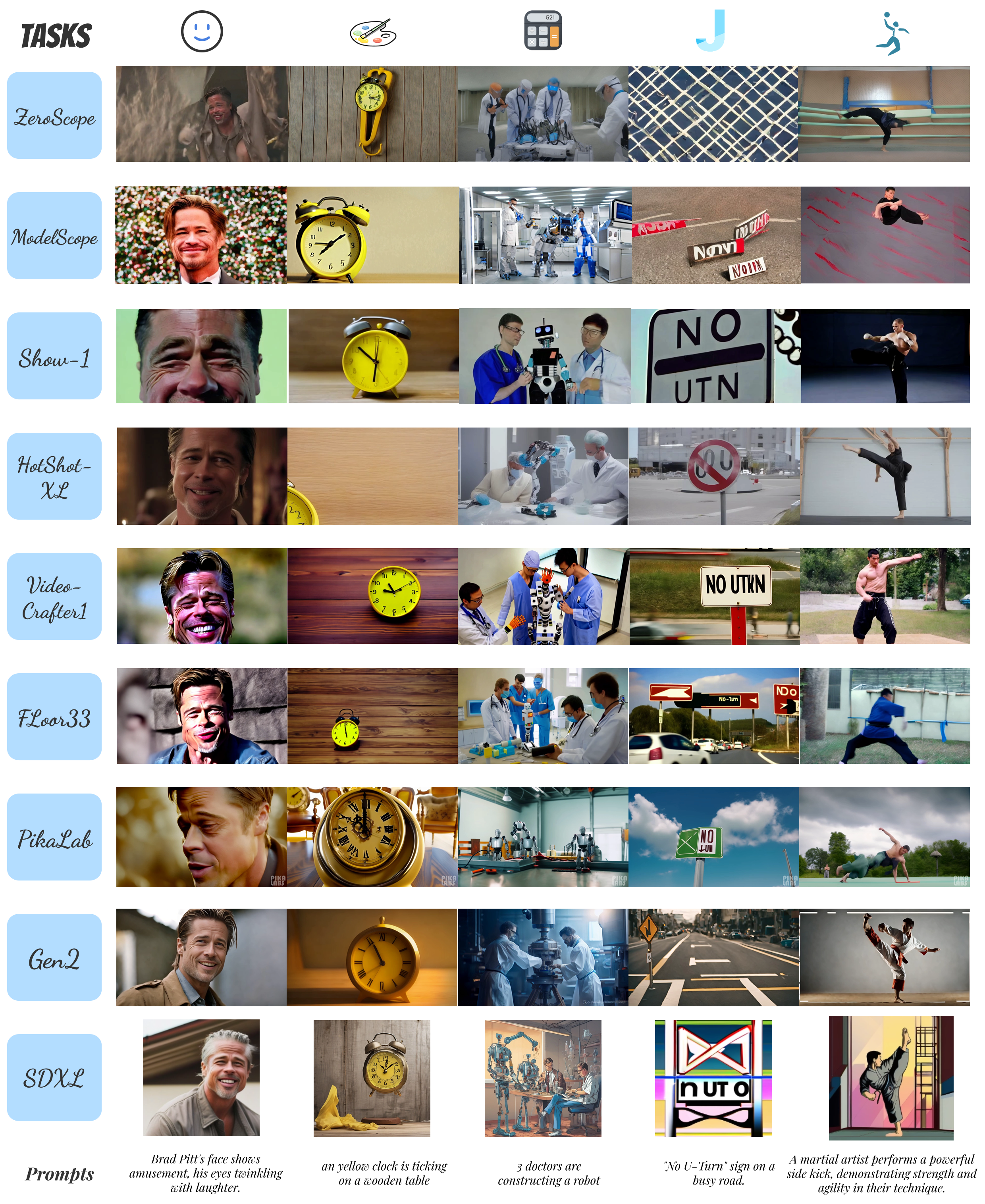}
    \vspace{-2em}
    \caption{Qualitative results of T2V models  in terms of different tasks (i.e., face generation, object  generation with color, object  generation with count, text generation, and activity generation)} 
    \label{fig:Vid_Gen_Task}
\end{figure*}

\end{appendices}

\fi

\end{document}